\documentclass[lettersize,journal]{IEEEtran}
\IEEEoverridecommandlockouts
\usepackage{cite}
\usepackage{amsmath,amssymb,amsfonts}
\usepackage{algorithmic}
\usepackage{graphicx}
\usepackage{textcomp}
\usepackage{colortbl}
\usepackage[dvipsnames]{xcolor}
\usepackage[font=footnotesize]{caption}
\usepackage{subcaption}
\usepackage[linesnumbered,ruled,vlined]{algorithm2e}
\usepackage[normalem]{ulem}
\usepackage{multirow}

\usepackage{tabularx}
\usepackage{tablefootnote}
\usepackage{threeparttable}
\usepackage{booktabs}

\usepackage[compact]{titlesec}
\titlespacing{\section}{0pt}{0.5ex}{0.5ex}
\titlespacing{\subsection}{0pt}{0.5ex}{0ex}
\titlespacing{\subsubsection}{0pt}{0.5ex}{0ex}

\SetAlCapNameFnt{\scriptsize}
\SetAlCapFnt{\scriptsize}

\def\BibTeX{{\rm B\kern-.05em{\sc i\kern-.025em b}\kern-.08em
    T\kern-.1667em\lower.7ex\hbox{E}\kern-.125emX}}
\begin{document}

\title{A Novel Implementation Methodology for Error Correction Codes on a Neuromorphic Architecture\\

\thanks{This work is partly supported by National Science Foundation (NSF) research project NSF CNS-1624668. We would like to thank the AMD Xilinx University Program for the continuous support through hardware development board donations. 
(Sahil Hassan and Parker Dattilo are Co-first authors.) (Corresponding author: Sahil Hassan.)\\
Sahil Hassan, Parker Dattilo, and Ali Akoglu  are with the Department of Electrical and Computer Engineering, University of Arizona, Tucson, AZ 85721, USA (email: sahilhassan@arizona.edu).}
}

\author{Sahil Hassan, Parker Dattilo, Ali Akoglu
\vspace{-3mm}
}

\maketitle

\definecolor{Gray}{gray}{0.9}

\begin{abstract}
The Internet of Things infrastructure connects a massive number of edge devices with an increasing demand for intelligent sensing and inferencing capability. 
Such data-sensitive functions necessitate energy-efficient and programmable implementations of Error Correction Codes (ECC) and decoders.  
The algorithmic flow of ECCs with concurrent accumulation and comparison types of operations are innately exploitable by neuromorphic architectures for energy efficient execution--an area that is relatively unexplored outside of machine learning applications. For the first time, we propose a methodology to map the hard-decision class of decoder algorithms on a neuromorphic architecture. We present the implementation of the Gallager B (GaB) decoding algorithm on a TrueNorth-inspired architecture that is emulated on the Xilinx Zynq ZCU102 MPSoC.  
Over this reference implementation, we propose architectural modifications at the neuron block level that 
result in a reduction of energy consumption by 31\% with a negligible increase in resource usage while achieving the same error correction performance.
\end{abstract}


\def \tabsize {0.85}

\section{Introduction} \label{sec:introduction}
Biologically inspired computing has been a growing area of research seeking energy efficient execution of complex tasks, beyond what is achievable by Von Neumann computers. Neuromorphic computing architectures are a realization of bio-mimicry in computing, with many applications designed to exploit the benefits in energy and parallelism~\cite{arXiv_2017_OtherPlatforms, PNAS_2016_Applications, IJCNN_2016_Applications, Rajendran2019lowpower, George2019IEEE}. Due to the event based execution inherent in a neural network model that is also foundational to neuromorphic architectures, machine learning with neural networks for classification and detection types of problems have been successfully deployed on them. Neuromorphic architecture research has been traditionally driven by improving accuracy for classification and detection types of problems in the trade space of scalability and energy efficiency ~\cite{TrueNorth,Davies2018Loihi,Neurogrid,Painkras2012SpiNNaker,Moradi2018DYNAPs,moreira_neuronflow_2020}. However, parallelism offered by neuromorphic architectures is a less explored area for non-traditional applications or deterministic algorithms~\cite{aimone_2022,Christensen22roadmap}. 

The rapidly growing number of connected IoT devices generate a large amount of multisource, highly heterogeneous data. Therefore, high-throughput and energy-efficient execution has become a key demand by applications of edge computing for distributed data-driven sensing, analysis, and inference. The networks used to connect IoT edge devices are characterized by resultant large streams of incomplete or incorrect data at the receiver due to channel interference and power limitations~\cite{herrero_2019}. This error in data can hamper the inference quality at edge devices and force them to retransmit the data. The retransmission process increases the communication cost and network traffic~\cite{matsui_2019}. Furthermore, repeated communication rapidly depletes the energy available to power constrained edge devices, and may lead to inaccurate inferences. Therefore in addition to the high-throughput and energy efficient execution, ensuring data integrity is also needed. The use of Error Correcting Codes (ECC) has been advocated for application in a power constrained IoT environment~\cite{alabady_2018}. 

To the best of our knowledge, neuromorphic architectures have not been explored for their potential on energy efficient error correction capability to support intelligent sensing operations at the edge. We believe that there is an opportunity for applying error correction algorithms to the neuromorphic domain due to two overlapping features: first in the computational model, and second in the execution flow. Among several classes of ECC, the Low-Density Parity Check (LDPC) codes are highly popular due to their channel capacity approaching performance~\cite{ldpc_capacity}. The key computations in an LDPC decoder algorithm such as accumulation and comparison are inherently supported by the neuron model~\cite{cassidy_cognitive_2013} in Spiking Neural Networks (SNNs) on modern neuromorphic architectures, and with slight modifications, can easily support modulo two addition. The iterative decoding process eventually terminates when a certain threshold has been reached for a set of parameters. This threshold  based operation is also naturally supported by the neuron model as it generates a spike only when a certain threshold has been reached. The execution flow of LDPC decoders offer concurrency and has already been the main target for parallelization ~\cite{unal_2018, unal_gabtestbed_2019, gronroos_ldpc_gpu_2012}. Such a parallel execution model is also innately supported by the axon-neuron connections of neuromorphic architectures. 

In this study we discuss mapping strategies to deploy an iterative decoding algorithm on neuromorphic architectures and propose architectural changes driven by the mapping strategies for efficient implementation. We demonstrate our methodology by implementing a representative decoder, with the aim of establishing a basis for extending to other decoders of the same class in the future. For this purpose, we chose the hard-decision Gallager B (GaB)~\cite{gallager_1962} decoding algorithm because its implementation contains foundational computing steps that are shared among other hard-decision error correction codes. We target a hard-decision class of algorithm as other classes of algorithms~\cite{unal_2018} involve complex arithmetic and floating point intensive operations making them not suitable for SNN-based implementation.
Therefore, to this end, the contributions of this paper are as follows:

\begin{itemize}
    \item Present the first mapping approach of an ECC decoder (GaB) on a neuromorphic architecture that is generalizable to other ECC decoders.
    \item Develop novel implementation of majority voting and, most resource efficient and deterministic multi-input XOR implementation on the neuromorphic architecture.
    \item Propose an XOR-integrated neuromorphic architecture that enables resource efficient and scalable implementations of relevant applications.
    \item Demonstrate throughput and energy benefits of the XOR-integrated ECC decoder implementation. 
\end{itemize}

We conduct our analysis by leveraging Reconfigurable Architecture for Neuromorphic Computing (RANC)~\cite{ranc}, an open-source highly-flexible ecosystem that consists of both software simulator and FPGA based emulator. RANC supports various neuromorphic architecture configurations through its parameterized design and has been validated for its ability to replicate TrueNorth behaviorally, which we use as a baseline implementation. For clarity of the illustrations, we present our mapping approach for GaB on the XOR-integrated neuromorphic architectures over a short LDPC code that generates 8-bit codeword. We utilize the cycle accurate software simulator to perform functional verification of the two implementations and quantify the cycle count reduction achieved with the XOR-integrated architecture over a test dataset. We emulate the baseline and XOR-integrated architectures on the Zynq UltraScale+ ZCU102 platform and execute GaB over the two emulations using the same test dataset. We show that our mapping approach on the XOR-integrated architecture reduces the cycle count to process a codeword by a factor of 1.48X with negligible increase of 1.29\% and 1.74\% in resource usage and power consumption respectively. We show that this negligible increase remains consistent by performing a scalability analysis where we compare resource usage and power consumption trends across designs with 15 and 25 neuromorphic cores. Finally we conduct a sweeping experiment where we vary the number of codewords tested and GaB simulation parameters, and observe that the XOR-integrated architecture saves up to 31\% energy over the baseline architecture. These results show that the proposed neuron block architecture offers execution time and energy benefits without sacrificing power efficiency for neuromorphic applications that rely on performing XOR operations.

The rest of the paper is organized as follows. Section~\ref{sec:background} provides background on the GaB algorithm. Section~\ref{sec:neuromorphic_model} describes the components and execution flow of the neuromorphic architecture. The neuromorphic mapping approach for fundamental functions frequently used in hard-decision decoders is presented in Section~\ref{sec:funcmap}. Section~\ref{sec:mapping} describes the mapping approach of GaB decoder. We present the experimental setup and implementation results in sections~\ref{sec:exp_setup} and~\ref{sec:results} respectively. Finally, section~\ref{sec:conclusion} concludes the findings of this paper.

\section{Background: Gallager B Decoding Algorithm} \label{sec:background}

\begin{figure}[t]
\vspace{-6mm}
    \centering
    \includegraphics[width=\linewidth]{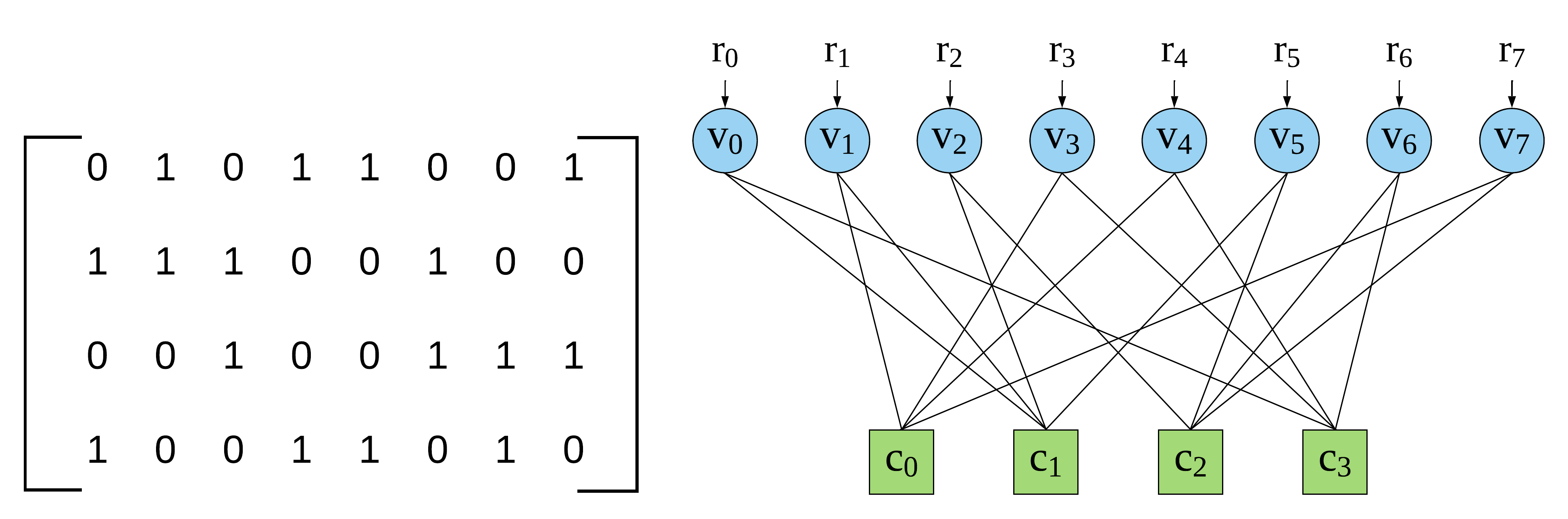}
    \caption{Running example H-matrix and corresponding Tanner graph.}
    \label{fig:hmat_tanner}
    \vspace{-6mm}
\end{figure}

GaB algorithm is representative of contemporary hard-decision class of error correction algorithms with its bipartite graph architecture, accumulation operations, and iterative decoding process~\cite{unal_2018,unal_gabtestbed_2019}. Therefore, as an entry point, implementing the GaB decoder on a neuromorphic computing architecture paves the way for establishing a methodology to implement sophisticated bit flipping algorithms and realize low power event driven error correction~\cite{ghaffari_2017} on the edge. In the following subsections, we introduce key terminology and present an algorithmic overview of GaB decoding. 

\subsection{Notations}
 LDPC codes are commonly described by a sparse matrix $H$ called parity-check matrix. The $H$ matrix dimensions are denoted as ($M,N$), where $N>M$. According to this code, a message $\mathbf{m}=(m_1, m_2,...,m_K)\in \{0,1\}^K$ (where, $K=N-M$) is encoded to generate a codeword $\mathbf{x}=(x_1, x_2,...,x_N)\in \{0,1\}^N$ with length $N$, which satisfies the condition $H\mathbf{x}^T = \mathbf{0}$. The decoder is designed as a bipartite Tanner graph, as depicted on the right of Figure~\ref{fig:hmat_tanner}, with $N$ Variable Node Units (VNUs) $v_n$ and $M$ Check Node Units (CNUs) $c_m$ ($1 \leq n \leq N$, $1 \leq m \leq M$), interconnected by edges according to the non-zero entries in the $H$ matrix. The number of edges connected to each node is called the degree of connection. The degree of VNU and CNU are denoted as $d_{v_n}$ and $d_{c_m}$ respectively. In this paper, we consider the regular LDPC code~\cite{gallager_1962}, which consists of all the VNUs having the same degree $d_v$ and all the CNUs having the same degree $d_c$. The set of CNUs connected with $v_n$ is expressed as $\mathcal{S}(v_n)$, and $\mathcal{S}(c_m)$ denotes the set of all VNUs connected with $c_m$. $v_nc_m$ indicates message passed from VNU $v_n$ to CNU $c_m$ and $c_mv_n$ indicates suggestions passed from CNU $c_m$ to VNU $v_n$.

\subsection{Gallager B Algorithm}
The Gallager B decoder is constructed as a bipartite Tanner graph according to the $H$ matrix. Figure~\ref{fig:hmat_tanner} shows an example of $H$ matrix and its corresponding Tanner graph. The circular nodes represent the VNUs and the square nodes represent the CNUs. At the beginning of the decoding process, each VNU receives a bit of $\mathbf{r}$, the message from the channel. The VNUs therefore broadcast these bits to their respective connected CNUs. The CNUs check for the parity condition between the bits received from connected VNUs, and send suggestions for corrections back to the VNUs, according to  Equation~\ref{eq:cnu}.

\footnotesize
\begin{equation}
\label{eq:cnu}
c_m v_n =  ( \sum_{t\in \mathcal{S} (c_m) \setminus v_n} v_t c_m) \mod 2
\end{equation}
\normalsize

\footnotesize
\begin{equation}
\label{eq:vnutocnu}
v_n c_m = \begin{cases} \mbox{0,} & \mbox{if {$(\sum_{t\in \mathcal{S} (v_n) \setminus c_m} c_t v_n) + r_n < b$}} \\
\mbox{1,} & \mbox{if {$(\sum_{t\in \mathcal{S} (v_n) \setminus c_m} c_t v_n) + r_n > b$}} \\
\mbox{$r_n,$} & \mbox{otherwise}
\end{cases}
\end{equation}
\normalsize

Upon receiving the suggestions from the connected CNUs, each VNU applies a majority voting function on the CNU suggestions and received $\mathbf{r}$ bit to calculate: (a) new estimated VNU-to-CNU signals according to Equation~\ref{eq:vnutocnu}, and (b) the estimated codeword bit according to Equation~\ref{eq:decide}. As illustrated with Algorithm~\ref{alg:gab}, a single decoding iteration involves computations based on Equations~\ref{eq:cnu},~\ref{eq:vnutocnu}, and~\ref{eq:decide}, which is repeated for a predefined number of iterations ($maxIter$ ).

\footnotesize
\begin{equation}
\label{eq:decide}
    x^\prime _n = 
    \begin{cases}
    \mbox{0,} & \mbox{if {$(\sum_{t\in \mathcal{S} (v_n)} c_t v_n) + r_n < b$}} \\
    \mbox{1,} & \mbox{if {$(\sum_{t\in \mathcal{S} (v_n)} c_t v_n) + r_n > b$}} \\
    \mbox{$r_n,$} & \mbox{otherwise}    
    \end{cases}
\end{equation}

\begin{equation}
\label{eq:syndrome}
    H\mathbf{x}^{\prime T} \mod 2= \mathbf{0}
\end{equation}
\normalsize
\setlength{\textfloatsep}{0pt}
At the end of each iteration, the estimated codeword $\mathbf{x'}$ is checked against the H-matrix to verify that it fulfills the parity conditions as captured by Equation~\ref{eq:syndrome}.
If $\mathbf{x^{\prime}}$ satisfies Equation~\ref{eq:syndrome} within the specified maximum number of iterations, then the codeword has been corrected successfully ($ \mathbf{x} = \mathbf{x^\prime}$). 
Despite reaching the maximum specified iteration, if Equation~\ref{eq:syndrome} is not satisfied, 
it is considered as a 'failed-to-decode' scenario.

\begin{algorithm}[t]
\scriptsize
\KwIn{$\mathbf{r}$, $H$, $maxIter$}
$ v_n c_m = r_n$ where $n \in \{1,2,...,N\}$ and $m \in \{1,2,...,M\}$\;
\For{$i \gets 1$ to $maxIter$}
{
    Compute all $c_m v_n$ using Equation~\ref{eq:cnu}\;
    Compute all $v_n c_m$ using Equation~\ref{eq:vnutocnu}\;
    Compute $ \mathbf{x^\prime}$ using Equation~\ref{eq:decide}\;    
    \If{Equation~\ref{eq:syndrome} holds true}
    {
        break\;
    }
}
\KwOut{$\mathbf{x^\prime}$}

\caption{GaB Decoding Algorithm}
\label{alg:gab}
\end{algorithm}
\setlength{\textfloatsep}{0pt}
\normalsize

\subsection{Architecture for 8-bit codeword}\label{subsec:8-bit_decoder}

\begin{figure}[b]
    \centering
    \includegraphics[width=0.8\linewidth, trim={0cm 0.5cm 0cm 0.5cm}, clip=true]{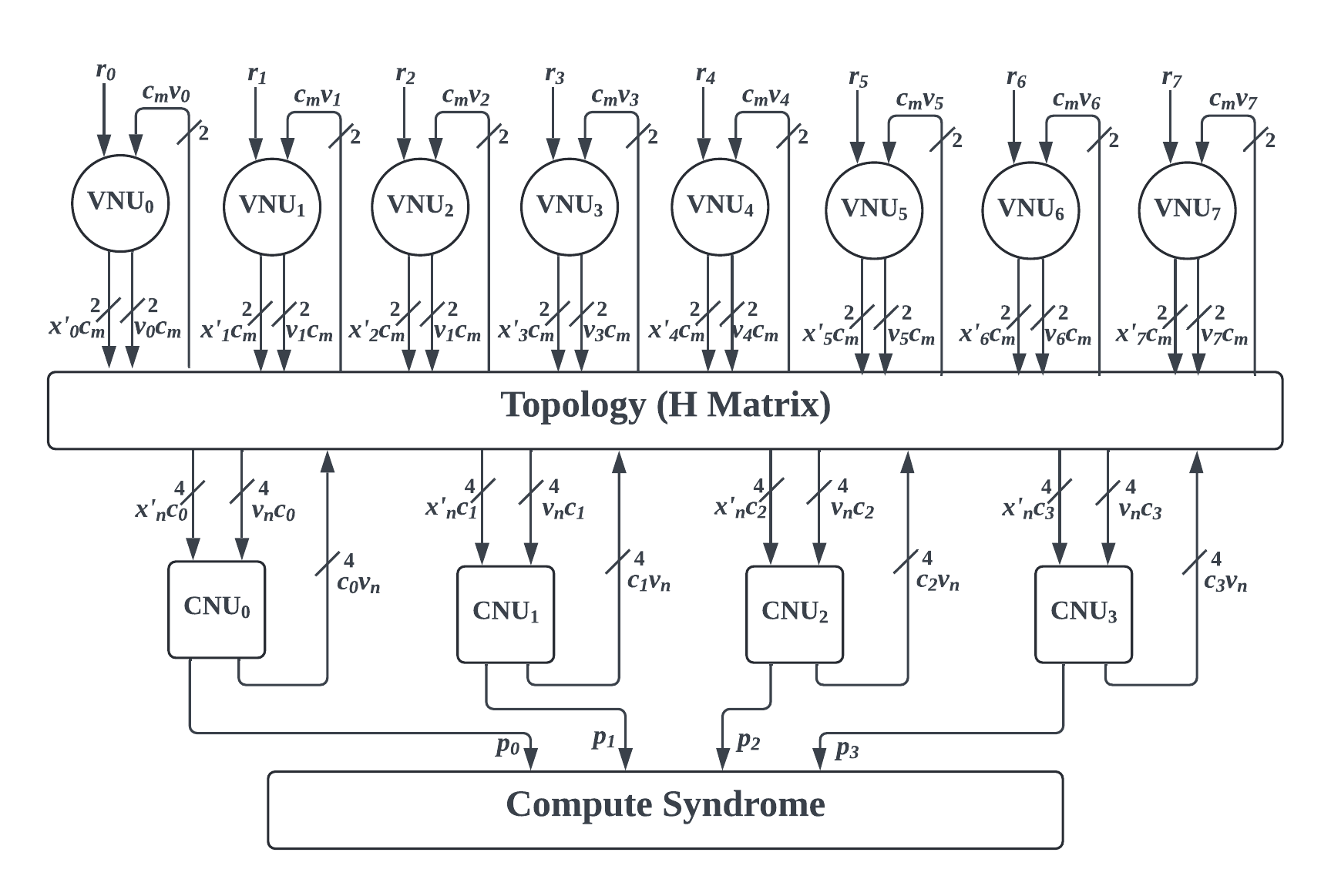}
    \caption{GaB Decoder architecture 8-bit codeword.}
    \label{fig:8bit_arch}
\end{figure}
In order to demonstrate mapping of a GaB decoder on the neuromorphic architecture, we consider an 8-bit LDPC code, and its decoder shown in Figure~\ref{fig:hmat_tanner} as a running example. The decoder parameters are $N$ = 8, $M$ = 4, $d_v$ = 2 and $d_c$ = 4. 
Figure~\ref{fig:8bit_arch} shows the fully unrolled architecture for this decoder where, each $VNU_n$ receives 2-bit $c_{m}v_{n}$ inputs from its connected CNUs and 1-bit of codeword form the channel ($r_n$). A VNU generates 2-bit $v_{n}c_{m}$ and 1-bit $x^\prime_{n}c_{m}$ based on Equation~\ref{eq:vnutocnu} and~\ref{eq:decide}, which is broadcasted to its two connected CNUs respectively. A $CNU_m$ receives 4-bit $v_{n}c_{m}$ and 4-bit codeword estimations $x^\prime_{n}c_m$ from the connected VNUs. Each CNU applies XOR operations as per Equation~\ref{eq:cnu} on the $v_nc_m$ signals to produce 4-bit $c_mv_n$ outputs. Furthermore, each CNU calculates a partial vector matrix multiplication (VMM) of Equation~\ref{eq:syndrome} (using XOR operations) for a given row of $H$, as $p_m$. Compute Syndrome unit receives the 4-bit $p_m$ signals generated by 4 CNUs and checks fulfillment of Equation~\ref{eq:syndrome}.

\section{Neuromorphic Model} \label{sec:neuromorphic_model}
\begin{figure}[t]
    \centering
    \includegraphics[width=0.8\linewidth]{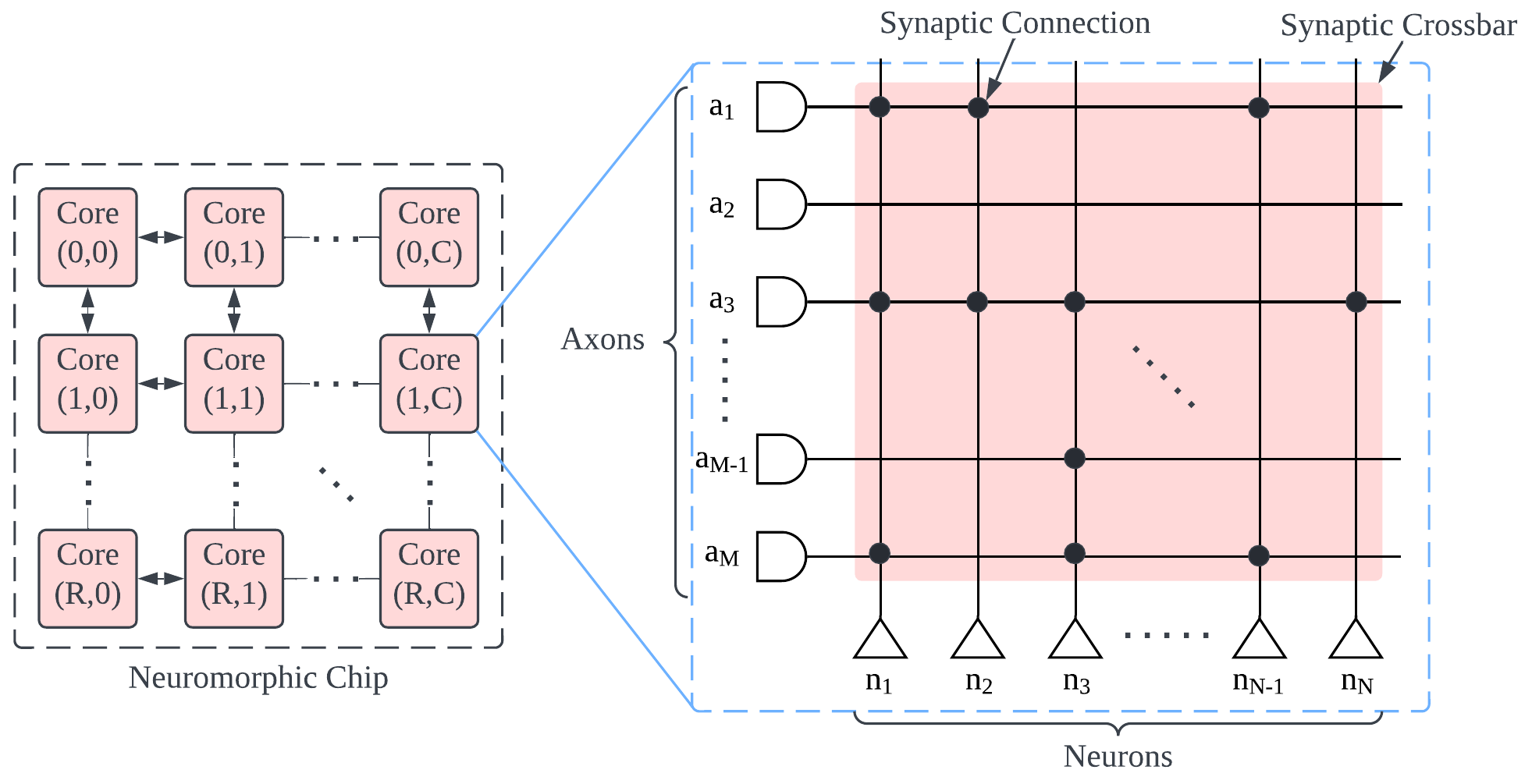}
    \caption{Logical representation of a neuromorphic core.}
    \label{fig:ranc_logical_bd}
\end{figure}

Figure~\ref{fig:ranc_logical_bd} shows the reference architecture composed of neuromorphic cores interconnected with a mesh network. The logical representation of a core is shown in Figure~\ref{fig:ranc_logical_bd} (right). The core consists of a set of $M$ axons (${a_{i}}$) and $N$ neurons (${n_{j}}$), which respectively receive the input data to the core and produce the output data from the core. The unit of these input-output data is called a \emph{spike} (\{0,1\} bit), that can be modeled in a time series as a Dirac delta function. The axons and neurons of a core are interconnected using crossbar connections called synaptic crossbar. The axon-neuron connection crosspoints on the synaptic crossbar can be configured to connect any desired set of axons to a given set of neurons. These connections are called synaptic connections, and are illustrated by a solid black circle at the connected crosspoints. Existence of a synaptic connection between an axon-neuron pair suggests that any spike received at the axon will be sent to the neuron. The inter-core activities at the design-level are synchronized by a global signal called a \emph{tick}. At each tick, a series of intra-core processing of input data received by axons takes place at neurons before moving on to the next tick.

\begin{figure}[t]
    \centering
    \vspace{-3mm}
    \includegraphics[width=0.85\linewidth, trim={0cm 0.5cm 0cm 0.5cm}, clip=true]{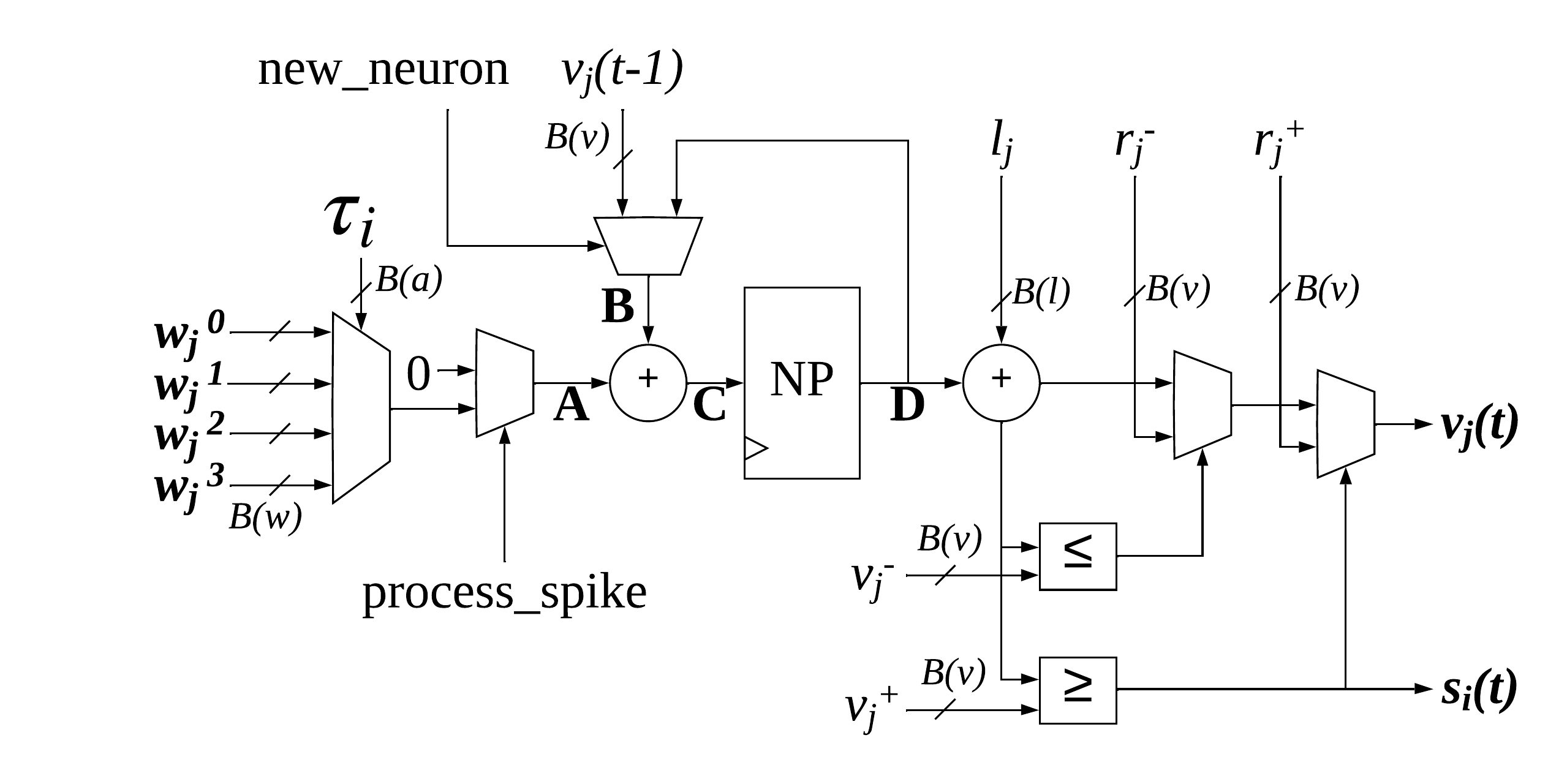}
    \caption{Neuron Block structure of the baseline architecture.}
    \label{fig:neuron_activecomp}
\end{figure}

\begin{table}[b]
\caption{Axon and Neuron Parameters. 
}
\vspace{-2mm}
\label{tab:axneu_param}
\tabcolsep=0.18cm
\centering
\scalebox{\tabsize}{
\begin{tabular}{|l|l|l|}
\hline
\rowcolor{Gray}
    Component & Static Parameter & Tick-specific Parameter\\ 
    \hline
    Axon $a_i$ & \begin{tabular}{c}Axon type $\tau_i$ \end{tabular} &- \\ 
    \hline
    \multirow{8}{*}{Neuron $n_j$} &
    \begin{tabular}{c}
        Weight array $\mathbf{w_{j}}$, $1{\le}j{\le}4$  \\
        Leak value $l_j$ \\
         Positive threshold $v_{j}^+$ \\
         Negative threshold $v_{j}^-$ \\
         Positive reset value $r_{j}^+$ \\
        Negative reset value $r_{j}^-$ \\
       Destination core \\
       offset $(\Delta x, \Delta y)$ \\
       Destination axon $a_d$
    \end{tabular} &
    \begin{tabular}{c}
        Current potential $v_{j}(t)$ \\
        Spike value $s_{j}(t)$ \\
        \\
        \\
        \\
        \\
    \end{tabular} \\
    
    \hline
\end{tabular}
}
\end{table}

Figure~\ref{fig:neuron_activecomp} illustrates the neuron block with architecture parameter definitions shown in Table~\ref{tab:axneu_param}. A given neuron $n_j$ (where $j \in \{1,2,..., N\}$) has an array of four weight values associated with it ($w_{j}[0]$ through $w_{j}[3]$). Similarly, each axon $a_i$ (where $i \in \{1,2,...,M\}$) has a specific type $\tau_i$ assigned to it out of four possible types ($\tau_i$ $\in$ \{0,1,2,3\}) that index into the weight array for each connected neuron. All of the neurons in each core follow the Leaky Integrate and Fire (LIF) neuron model. Algorithm~\ref{alg:neuron} shows the overall execution of a neuron and  this process is repeated during each tick concurrently by all $N$ neurons of a core, before moving on to the next tick.

\setlength{\textfloatsep}{2pt}
\begin{algorithm}[t]
    \scriptsize
    \KwIn{$\mathbf{w_j}$, $v_{j}^+$, $v_{j}^-$, $l_j$, $r_{j}^+$, $r_{j}^-$, $maxTicks$}
    $t = 0$;\\
    $v_{j}(0)$ = 0;\\
    \For{$t \gets 1$ to $maxTicks$}
    {
        \For{$j \gets 1$ to $N$} {
            $s_{j}(t) = 0$;\\
            $v_{j}(t) = v_{j}(t-1)$;\\ 
            \For {$a_i$ in Connected axons \& $a_i$ Spiking}
            {
                $v_{j}(t) += w_{j}[\tau_i]$;
            }
            $v_{j}(t)$ += $l_j$;\\
            \If{$v_{j}(t) \geq v_{j}^+$}
            {
                $s_{j}(t) = 1$;\\
                $v_{j}(t) = r_{j}^+$;
            }
            \If{$v_{j}(t) \le v_{j}^-$}
            {
                $v_{j}(t) = r_{j}^-$;
            }
    
            Send $s_{j}(t)$ to destination core;\\
        }
    }
\caption{Neuron $n_j$ operation}
\label{alg:neuron}
\end{algorithm}
\normalsize

The operation of the neuron $n_j$ starts by observing the first axon on the core. If there is a synaptic connection and a spike is present on the axon $a_i$, the \emph{process\_spike} signal will select the weight value $w_{i,j}$ based on the type $\tau_i$ of the axon. This weight value in the neuron block is labeled as operand $A$ in Figure~\ref{fig:neuron_activecomp}. In the case that either no synaptic connection exists or a spike is not received on a connected axon, the \emph{process\_spike} signal will select 0 as operand $A$. The second operand $B$ is obtained by loading remaining neuron potential from the previous tick $v_j(t-1)$ when the neuron block starts processing for $n_j$ neuron (\emph{new\_neuron} = 1). Accumulation of $A$ over the previous tick potential ($B$) produces $C$, which gets stored as neuron potential through \emph{NP} register. The feedback loop is used to keep accumulating the neuron potential for the remaining axons (\emph{new\_neuron} = 0). Upon completing the processing of all axons, the leak $l_j$ is accumulated with $D$, and resulting neuron potential is compared against $v_j^+$ and $v_j^-$. If neuron potential is greater than equal to the positive threshold $v_j^+$, a spike is produced as $s_j(t)$, and the neuron potential $v_j(t)$ is set to $r_j^+$. Otherwise, the comparator outputs no spike, and the current value of neuron potential is assigned to $v_j(t)$.

\section{Function Mapping on RANC}
\label{sec:funcmap}
In this section, we identify the functions that are frequently found in the execution of the GaB decoding algorithm, and present our approach for mapping them on the neuromorphic platform. Based on Algorithm~\ref{alg:gab} and Equations~\ref{eq:cnu}-\ref{eq:syndrome}, the architecture needs to  support register or buffer, majority, AND, OR, NOR and XOR functions. 
XOR function is a special case due to its linearly inseparable nature therefore we start with mapping the 4-input XOR that is utilized in the CNU block.

\subsection{XOR} \label{subsec:xor}
XOR is a key operation used frequently in application domains such as ECC and cryptography. Equations~\ref{eq:cnu} and~\ref{eq:syndrome} of the GaB decoding algorithm presented in Section~\ref{sec:background} show the use of XOR in the form of modulo 2 additions. Due to the extensive use of this function, there have been several efforts in literature to mimic XOR behavior using neural networks~\cite{delaney_2017}. These works acknowledge that the XOR is a linearly inseparable problem, and unsolvable using a single real-valued neuron. Therefore, the existing mapping of the 2 input XOR function on neuromorphic platforms~\cite{cassidy_cognitive_2013} show that XOR implementation requires 3 LIF neurons, which is much higher compared to the remaining logical operations (AND, NAND, OR, NOR, NOT) which require only 1 neuron. Furthermore, these three neurons operate over two ticks to produce the output of a single 2-input XOR operation, which leads to increased execution time and routing activity.

\begin{figure}[t]
    \centering
    \includegraphics[width=0.9\linewidth]{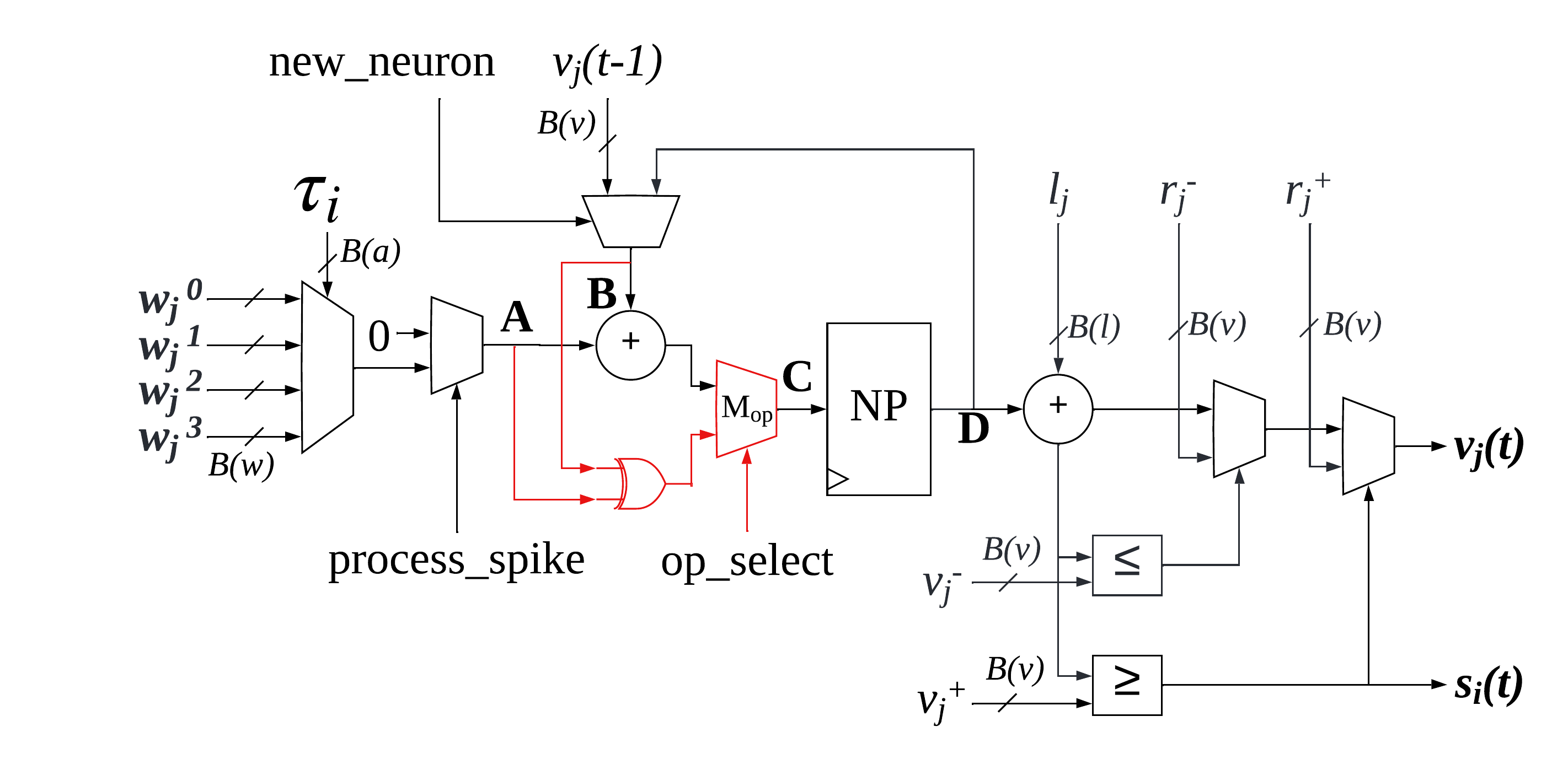}
    \vspace{-2mm}
    \caption{XOR integrated Neuron Block architecture.}
    \vspace{-2mm}
    \label{fig:neuronblock_xor}
\end{figure}

\begin{figure}[t]
    \centering
    \includegraphics[width=0.65\linewidth, trim={0cm 0.2cm 0cm 0.5cm}, clip=true]{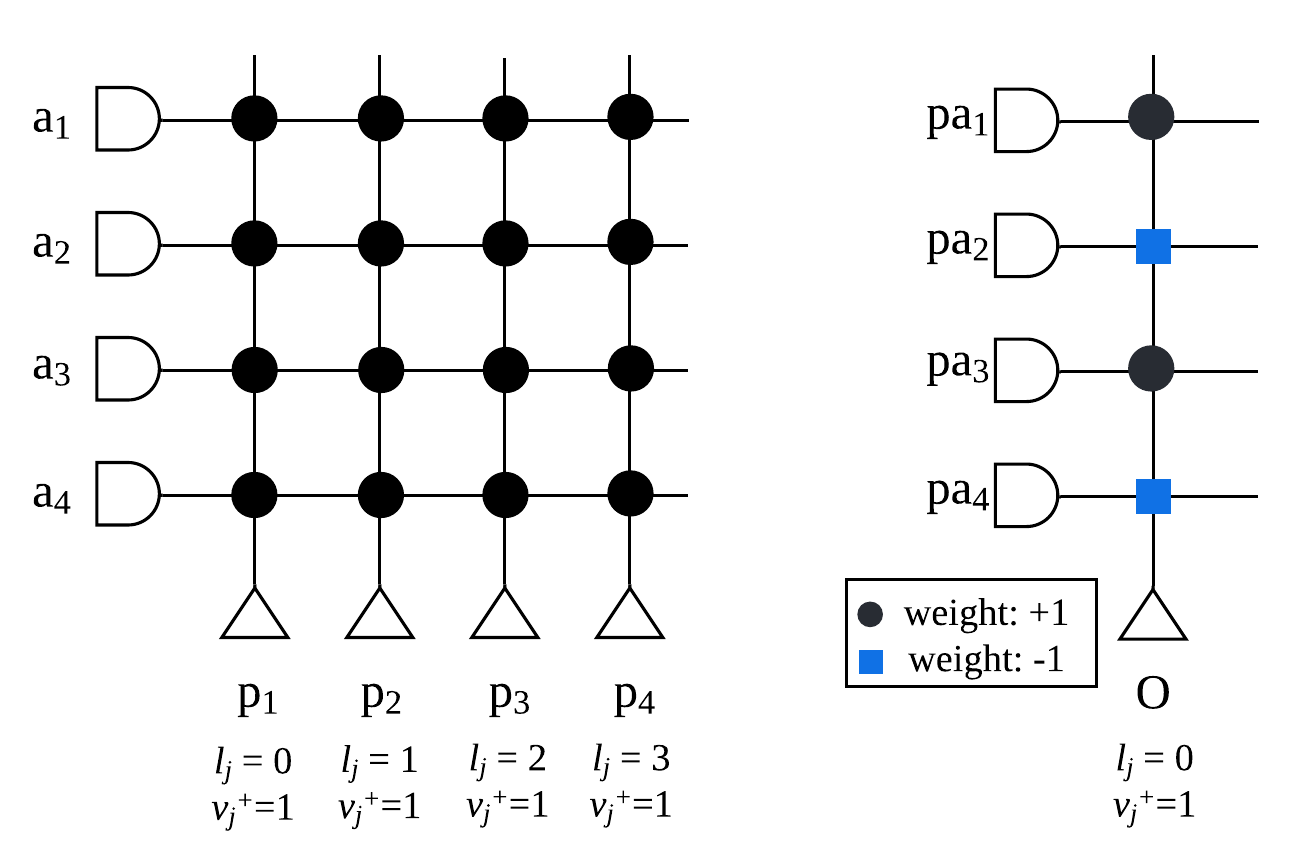}
    \caption{Implementation of 4-input XOR function on baseline architecture.}
    \label{fig:xor_tn}
\end{figure}

\begin{table}[b]
    \centering
    \vspace{1mm}
    \caption{4-input XOR mapping ($a_0-a_3$ represented in hexadecimal form).}
    \vspace{-2mm}
    \scalebox{\tabsize}{
    \begin{tabular}{|c|c|c|}
        \hline
        \rowcolor{Gray}
         \{$a_1$ $a_2$ $a_3$ $a_4$\} (HEX)& $p_1$ $p_2$ $p_3$ $p_4$ & $O$\\ \hline
         0 & 0 0 0 0 & 0 \\ \hline
         1, 2, 4, 8 & 1 0 0 0 & 1 \\ \hline
         3, 5, 6, 9, A, C & 1 1 0 0 & 0 \\ \hline
         7, B, D, E & 1 1 1 0 & 1 \\ \hline
         F & 1 1 1 1 & 0 \\ \hline
    \end{tabular}
    }

    \label{tab:xor_truthtable}
\end{table}

Motivated by the higher neuron requirement of the XOR implementation posing as a resource scarcity issue, we modify the neuron block architecture in RANC as shown in Figure~\ref{fig:neuronblock_xor}, such that a single neuron can perform an XOR operation only on the least significant bit of the two inputs A and B, and sets all the upper bits to 0. The neuron block has an additional single bit static parameter for operation selection ($op\_select$), that chooses between LIF and XOR behavior. This $op\_select$ bit is used as the select input to the multiplexer $M_{op}$, that picks between the outputs from the adder and the XOR module, to forward as the output of the neuron block. With this neuron block modification, each XOR operation can be implemented with only one neuron in a single tick. For the rest of this paper, we refer to this XOR based modified neuron block design as XOR integrated architecture, and the LIF based unmodified neuron block design from Figure~\ref{fig:neuron_activecomp} as baseline architecture.

To demonstrate the resource efficiency benefit of the presented modified neuron block architecture, we establish a baseline comparison point with a 4-input XOR implementation in Figure~\ref{fig:xor_tn}, along with its corresponding truth-table in Table~\ref{tab:xor_truthtable}. This implementation requires two layers of neurons, operating over two ticks. The first layer in Figure~\ref{fig:xor_tn} (left) counts the number of spikes (1's) present in the input signal received in any order. For example, if three spikes are received at $a_1$-$a_4$, the neurons $p_1$, $p_2$ and $p_3$ produce a spike. This behavior is achieved by using a fully connected core with a single type of axons indexing to a weight value of 1. Each neuron of the core performs weight accumulation with a positive threshold of 1, a hard-reset that sets the potential back to 0 after a spike, or if the potential is below 0, and a leak value that starts with 0 for $p_1$ neuron, and keeps increasing by 1 for the remaining neurons (leak values 1, 2, 3 for $p_2$, $p_3$, $p_4$ respectively). The spikes produced by the first layer neurons are sent to the second layer illustrated in Figure~\ref{fig:xor_tn} (right). The second layer has only one neuron that is fully connected with all the axons. The axons are of two types that index into two weight values: \{1, -1\}. The $pa_{odd}$ axons index to a weight of 1, and $pa_{even}$ axons index to a weight of -1. The $O$ neuron performs accumulation with a positive threshold of 1, and a leak value of 0. The design of the second core/layer implies that neuron $O$ will only spike when an odd number of 1s is propagated through the previous core. The truth table for the presented XOR mapping is illustrated in Table~\ref{tab:xor_truthtable}. With this presented mapping methodology, an $n$-input XOR operation would require $n+1$ neurons, $n$ neurons in the first layer, and 1 in the second layer. Therefore, for a 2-input XOR operation, 3 neurons are required, which is in line with the presented neuron requirement in~\cite{cassidy_cognitive_2013}. There are several existing 2-input XOR implementations in the literature using Spiking Neural Networks (SNNs)~\cite{delaney_2017, zeigler_2017, cyr_2020, matsumoto_2018, lingfei_2021, enriquez_2018}. These works are limited due to their lack of consideration of multi-input XOR operations. Furthermore, the presented approaches in~\cite{delaney_2017, zeigler_2017, matsumoto_2018, lingfei_2021} require more number of neurons compared to the presented baseline approach. There are additional complexities of a training process of the SNNs~\cite{delaney_2017, matsumoto_2018, lingfei_2021, enriquez_2018}, and requirement of maintaining specific temporal sequence between arrival of specific signals~\cite{zeigler_2017}. To the best of our knowledge, the presented baseline XOR approach is the most resource efficient, and deterministic implementation of multi-input XOR operation using the LIF neuron model.

\vspace{-0.35cm}
\footnotesize
\begin{align}
    N_{Neurons} & = (M \times d_c \times d_c) + (M \times (d_c + 1)) \nonumber \\
    & =  M \times ({d_c}^{2} + d_c + 1)  \nonumber \\
    & =   \frac{N}{2} \times ({d_c}^{2} + d_c + 1)
  \label{eq:xor_neu_count}
\end{align}
\vspace{-0.5cm}
\begin{align}
    N_{Mod-Neurons} & = (M \times d_c \times 1) + (M \times 1) \nonumber \\
    & =  M \times (d_c + 1) = \frac{N}{2} \times (d_c + 1) 
  \label{eq:opt_xor_neu_count}
\end{align}
\normalsize

In order to quantitatively analyze the benefit of the XOR integrated neuron block implementation in a real-world application, let us consider a GaB decoder implementation described by $N$, $M$, $d_v$ and $d_c$ (given $M$ = $\frac{N}{2}$). This design consists of $M$ CNUs, each performing $d_c$ number of ($d_{c} -1$)-input XORs and one $d_c$-input XOR. Equation~\ref{eq:xor_neu_count} expresses the number of required LIF neurons for XOR operations ($N_{Neurons}$) and Equation~\ref{eq:opt_xor_neu_count} presents the number of XOR integrated neurons ($N_{Mod-Neurons}$) required for XOR operations in the same decoder. Therefore, the 8-bit decoder presented in Section~\ref{subsec:8-bit_decoder} requires 84 LIF neurons, while only 20 XOR integrated neurons are required for the same decoder. Both Equations~\ref{eq:xor_neu_count} and~\ref{eq:opt_xor_neu_count} show that the number of required neurons for XOR is directly proportional to $d_c$ and $N$ values. 

\begin{figure}[t]
    \centering
    \includegraphics[width=0.9\linewidth]{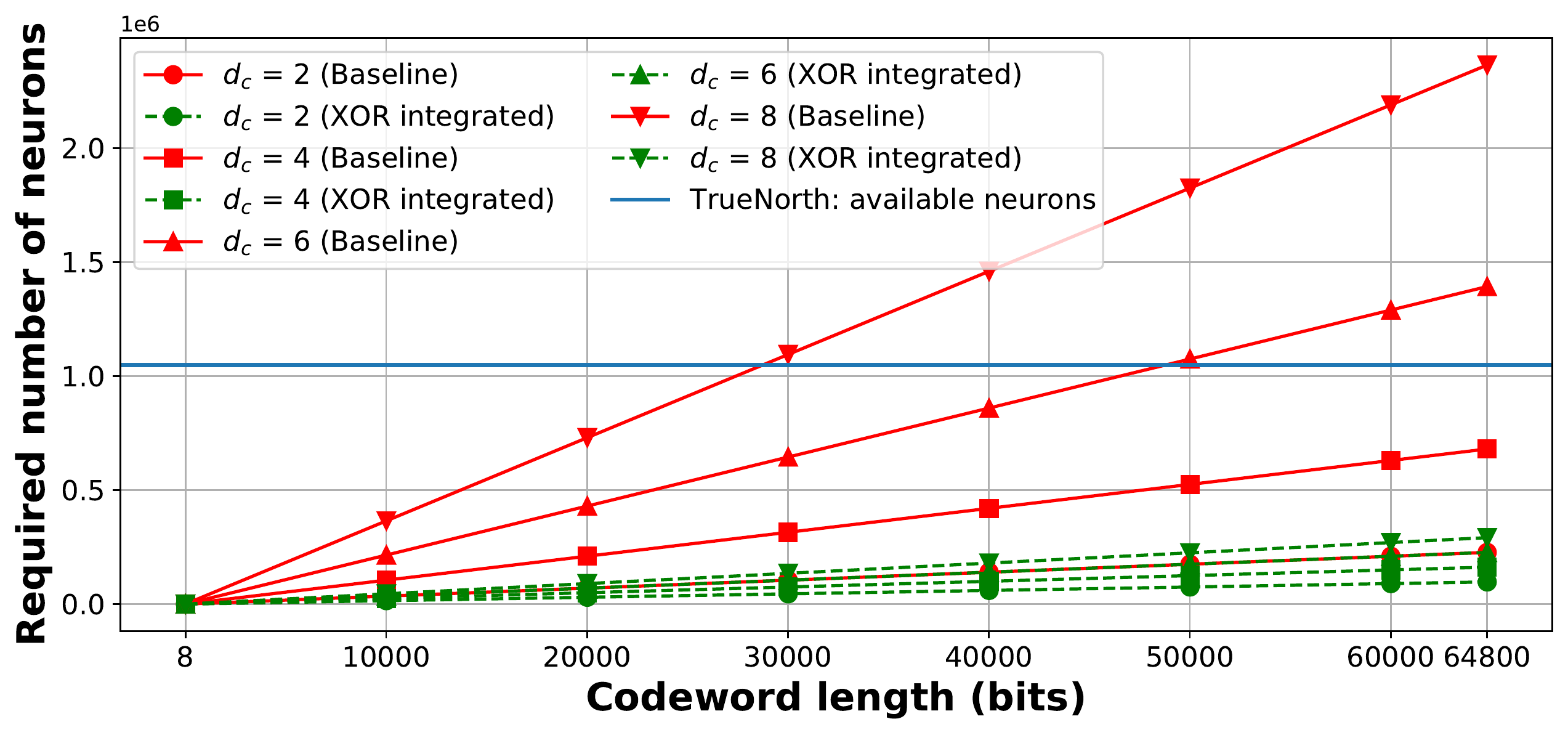}
    \caption{Required neurons to map XOR operations with respect to $N$ and $d_c$ on baseline and XOR integrated architectures on RANC.}
    \label{fig:ldpcxorgraph}
\end{figure}

In practical implementations of LDPC codes, $d_c$ varies from 2 to 8, and $N$ can grow to 64,800 bits in Digital Video Broadcasting-S2~\cite{unal_2018,ghaffari_2017}. Figure~\ref{fig:ldpcxorgraph} shows how the number of required neurons scales when mapping XOR operations with respect to $N$ and $d_c$ values for the baseline and XOR integrated architectures using Equations~\ref{eq:xor_neu_count} and~\ref{eq:opt_xor_neu_count}. The horizontal line marks the number of available neurons on the TrueNorth chip. The four solid and dashed line plots correspond to LIF neurons and XOR modified neurons respectively over four $d_c$ values, \{2, 4, 6, 8\}. As $N$ approaches 64,800, the XOR operations alone occupy more than half of the available neurons on the TrueNorth chip when $d_c$ is 4. For $d_c$ values of 6 and 8, the XOR operations occupy all the neurons when $N$ reaches 48,770 and 28,728 bits respectively. Note, the largest mappable $N$ is significantly lower than these two values, as the underlying decoder functions besides XORs, require resources as well. Therefore, for $d_c$ values of 6 and 8, decoders with only smaller sized LDPC codes are realizable on the neuromorphic platforms using baseline LIF neurons. In contrast, number of required neurons with the XOR integrated neuron block is well within the neurons present on the platform. For the largest $d_c$ and $N$ values of 8 and 64,800 respectively, the XOR operations of GaB would require only 291,600 neurons, which is 27.8\% of the total available neurons. In Section~\ref{sec:mapping} we present the detailed GaB decoder design on the neuromorphic core with the XOR integrated neuron block. Then, in Section~\ref{sec:results}, we discuss power and energy differences between the GaB decoder design utilizing the XOR integrated neuron versus the design utilizing the baseline LIF neuron.

\subsection{Register or Buffer}
\begin{figure}[t]
    \centering
    \includegraphics[width=0.8\linewidth]{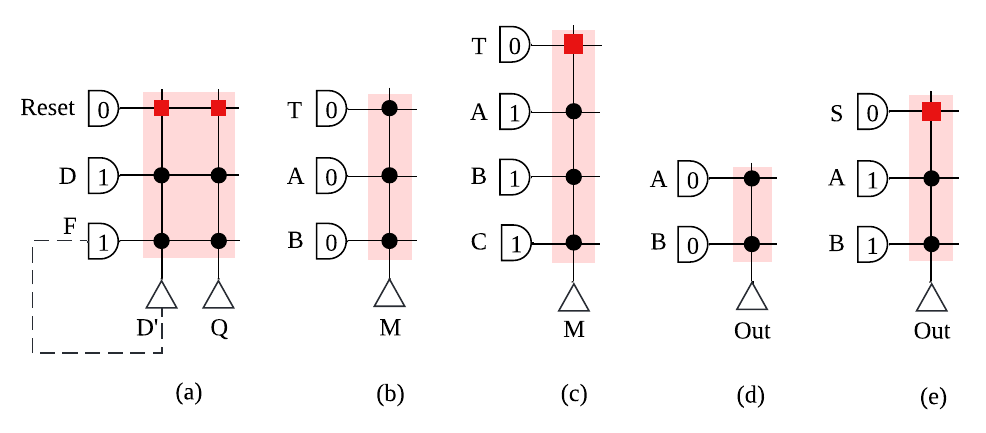}
    \caption{Function mapping for: (a) Register/buffer, (b) 3-input Majority and (c) 4-input Majority, (d) 2-input AND and OR, (e) 2-input NOR.}
    \label{fig:funcMapRanc}
\end{figure}

We design a 1-bit register/buffer as  shown in Figure~\ref{fig:funcMapRanc}(a) where $D$ and $Q$ are the input and output bits respectively. The $Q$ neuron has a positive threshold of 1, and the $D$ axon indexes into a weight value of 1 for the $Q$ neuron. Through this configuration, any spike received on $D$ generates a spike on $Q$ neuron. We introduce a feedback loop to preserve a received spike as a memory on $Q$, so that it keeps spiking until the register is reset. Through the axon $F$ and neuron $D'$ feedback loop, when $D$ axon receives a spike, $D'$ also spikes where $D$ indexes into weight of 1 for $D'$ neuron, which has a positive threshold of 1. $D'$ feeds the spike back to the $F$ axon in the next tick, creating a feedback loop. $F$ having a synaptic connection with the $Q$ neuron, will cause the $Q$ neuron to spike. Hence, for the subsequent ticks, even if there are no spikes received on $D$, the feedback loop with $F$ axon and $D'$ neuron helps preserve the spike on $Q$ neuron.

We introduce the $Reset$ axon to set the register value to 0, where $Reset$ has synaptic connections with $D'$ and $Q$ neurons, and it indexes into a negative weight value of -2. This negative weight should be large enough in magnitude to drive the neuron potential of $D'$ and $Q$ to $\leq 0$, so that they don't spike anymore in subsequent ticks, which is equivalent to preserving a value of 0. We use the proposed model for a 1-bit register to implement an n-bit register/buffer by cascading each register sequentially, scaling up to a 256-bit register within a single core using a single $Reset$ signal. An example multi-bit register/buffer will be shown in section~\ref{subsec:input_core_mapping}.

\subsection{Majority Voting}
We implement two versions of a majority voting function as shown in Figure~\ref{fig:funcMapRanc}(b) and (c). This design mimics/emulates the function described by Equation~\ref{eq:vnutocnu}. There are three possible states that the majority function applied over the received multi-bit inputs. These states are- majority of ones, majority zeros and tie between the two. The design choices for mapping majority voting for odd and even number of inputs vary as illustrated with the implementations shown in Figures~\ref{fig:funcMapRanc}(b) and (c) respectively. Both designs have a single output neuron $M$, which has synaptic connection with all the input axons. In case of the odd input count design, all the axons index into the same weight of 1. As there can only be a majority case and no tie case, all the axons are weighted equally. If more than half of the axons \{$T$, $A$, $B$\} (any 2 or 3) spike, $M$ should also spike. Hence a neuron potential of $\geq 2$ before applying leak and threshold, should cause $M$ to spike in this case. We realize this by setting the positive threshold $v^+ = 1$ and leak $l = -1$ for the $M$ neuron. Leak is used here to ensure no remainder potential from the current tick moves on to the next tick.

In case of the even input count design with 4 axons (\{$T$, $A$, $B$, $C$\}), a tie case is also possible besides the majority cases. The occurrence of a tie should cause the $M$ neuron to take the value according to the $T$ axon only. This axon is the tie input of the function. Hence, axon $T$ is weighted higher at weight value of 2, and remaining axons index into weight value of 1. This one higher weight assignment helps break the tie based on value at $T$ axon. This function would achieve a majority case, if at least 3 out of 4 axons have the same value, which corresponds to a neuron potential of $\geq 3$. Hence a positive threshold $v^+ = 1$ and leak $l = -2$ is chosen for $M$ neuron. In tie cases, where only two out of four axons receive spike, the neuron potential can only reach 3 when $T$ is one of the spiking axons. Hence, $M$ neuron would only spike if $T$ axon has a spike on it.

\subsection{Boolean operations} \label{subsec:boolean}
GaB utilizes AND, OR, and NOR operations to check conditions for terminating the decoding process and generating the output codeword when all of the CNUs have their parity conditions fulfilled (Equation~\ref{eq:syndrome}). We implement 2-input AND and OR operations as illustrated in Figure~\ref{fig:funcMapRanc}(d), where both $A$ and $B$ input axons are connected with $Out$ neuron, and index to a weight of 1. In case of AND operation, $Out$ should only generate a spike when spikes are received on both $A$ and $B$, driving the neuron potential in $Out$ to 2, before applying any leak or threshold. Hence, we choose $v^+ = 1$ and $l = -1$. In case of OR however, reaching neuron potential of 1 before applying leak is sufficient to capture the condition and generate a spike, hence, $v^+ = 1$ and $l=0$ is chosen for the $Out$ neuron.

We implement 2-input NOR operation as illustrated in Figure~\ref{fig:funcMapRanc}(e). This mapping has an additional $S$ axon besides the $A$ and $B$ axons carrying input operands. The $S$ axon is defaulted to receive a spike. Presence of spike in either of the two input axons ($A$ and $B$) should stop $Out$ neuron from spiking. Hence, we associate a negative weight of -1 with $A$ and $B$ axons to penalize the neuron potential when $A$ or $B$ has a spike. $S$ axon indexes into a positive weight of 1. As a result, the neuron potential of $Out$ can reach to 1 (without applying leak or threshold), only when both $A$ and $B$ axons do not spike. Therefore, we set positive threshold $v^+ = 1$ and leak $l = 0$, with reset value $r^+ = 0$ for $Out$ neuron.

Note that all of the above mentioned boolean function mapping approaches provide competitive resource utilization with existing literature~\cite{cassidy_cognitive_2013}, where only one neuron has been used to map AND, OR and NAND operations.

\section{Mapping Methodology: GaB on RANC}
\label{sec:mapping}
\begin{figure}[t]
    \centering
    \includegraphics[width=0.9\linewidth]{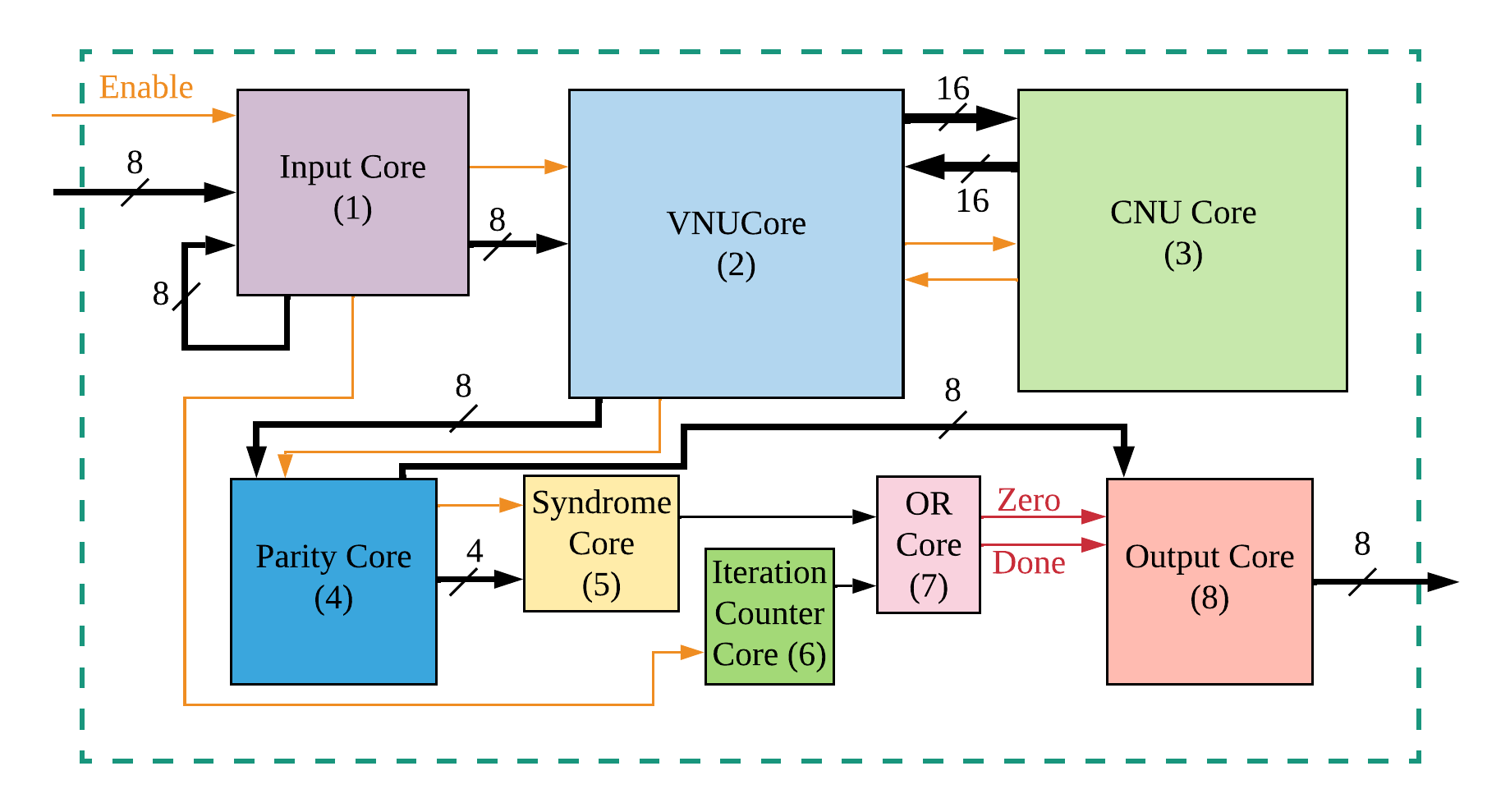}
    \vspace{-2mm}
    \caption{Overview of GaB algorithm on a neuromorphic architecture using the proposed XOR-integrated neuron block.}
    \label{fig:overalldesign}
\end{figure}

Figure~\ref{fig:overalldesign} shows the hardware implementation of GaB that utilize the functions listed in Table~\ref{tab:functionlist} and the XOR integrated neuron block. In the following subsections  we present our detailed mapping approach for each of the 8 cores, and discuss their functionality through tick-by-tick execution of a single GaB iteration based on our running example with the selected input $\mathbf{r} = 10001100$ that is corrected to  $10001101$.

\begin{table}[t]
\vspace{-3mm}
\caption{List of cores and associated functions.}
\vspace{-2mm}
\label{tab:functionlist}
\tabcolsep=0.18cm
\centering
\scalebox{0.8}{
\begin{tabular}{|l|l|}
\hline
\textbf{Core Name} & \textbf{Associated Functions} \\ \hline
Input & Register, Buffer \\ \hline
VNU & 2 and 3-input Majority Voting \\ \hline
CNU & 3-input XOR ($mod$ $2$ addition) \\ \hline
Iteration Counter & Counter, Comparator \\ \hline
Parity  & 4-input XOR ($mod$ $2$ addition) \\ \hline
Syndrome & 4-input NOR \\ \hline
OR & 2-input OR \\ \hline
Output & 2-input AND \\ \hline
\end{tabular}
}
\end{table}

\subsection{Input Core}\label{subsec:input_core_mapping}
\begin{figure}[t]
    \centering
    \includegraphics[width=0.7\linewidth]{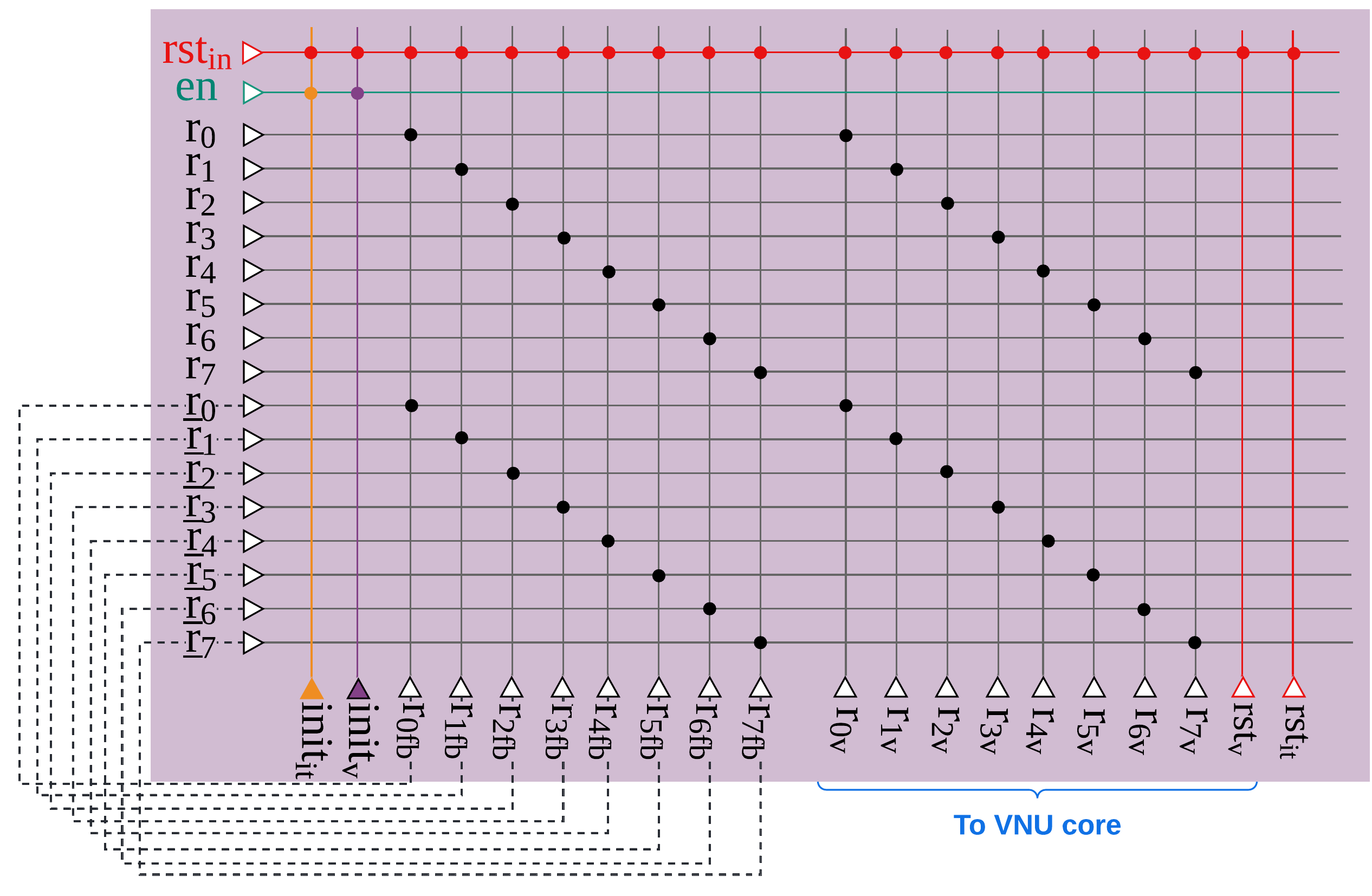}
    \caption{Input Core.}
    \label{fig:inputcore}
\end{figure}

The \emph{Input Core} shown in Figure~\ref{fig:inputcore} receives 8-bits of frame data $\mathbf{r}$ through input axons $r_0 - r_7$, where axons are indexed from top to bottom. The feedback loop retains $\mathbf{r}$ through axons \underbar{$r_0$}$-$\underbar{$r_7$}. The $en$ axon is used to provide spike for initiating the decoding process. The axons of this core are of the same type with the weight $w \in \{1\}$. The neurons of this core have a threshold value ($v_{j}^+$) of 1 and reset value ($r_{j}^+$) of 0 as shown in Table~\ref{tab:input_axneuconfig}. In the layout shown in Figure~\ref{fig:inputcore}, neurons are indexed from left to right in increasing order. This core utilizes 20 neurons where, 8 neurons $r_{0_{fb}}-r_{7_{fb}}$ function as an 8-bit register for the input frame from axons $r_0 - r_7$ and provide feedback to retain $\mathbf{r}$, 8 neurons $r_{0_{v}}-r_{7_{v}}$ forward data to the \emph{VNU Core} and the remaining 4 neurons $init_{it}, rst_{it}, init_v,$ and $rst_v$ are control signals for initializing and resetting the \emph{Iteration Counter} and \emph{VNU} Cores respectively.

\begin{table}
\vspace{-3mm}
\caption{Axon and Neuron configurations of \emph{Input Core}.}
\vspace{-2mm}
\label{tab:input_axneuconfig}
\tabcolsep=0.18cm
\centering
\scalebox{\tabsize}{
\begin{tabular}{|l|l|l|l|l|l|l|}
\hline
\rowcolor{Gray}
\multicolumn{7}{|c|}{Axons} \\ \hline
\multicolumn{5}{|l}{Name} & \multicolumn{2}{|l|}{Type $\tau$} \\ \hline
\multicolumn{5}{|l}{$en$ - \underbar{$r_7$}} & \multicolumn{2}{|l|}{0}  \\ \hline
\multicolumn{5}{|l}{$rst_{in}$} & \multicolumn{2}{|l|}{1} \\ \hline
\rowcolor{Gray}
\multicolumn{7}{|c|}{Neurons} \\ \hline
Name & $v^+_j$ & $l_j$ & $r^+_j$ & Rst type & Op Type & $w_j$ \\ \hline
$init_{it}$, $init_v, r_{0_{fb}}$-$r_{7_v}$ & 1 & 0 & 0 & Hard & 0 & $[1,-1,1,1]$\\ \hline
$rst_{it}, rst_v$ & 1 & 0 & 0 & Hard & 0 & $[1,1,1,1]$\\ \hline
\end{tabular}
}
\end{table}

\begin{table}
\caption{\emph{Input Core} neuron activity for $10001100$ input word over 2 ticks.}
\vspace{-2mm}
\label{tab:inputneurons}
\tabcolsep=0.18cm
\centering
\scalebox{\tabsize}{
\begin{tabular}{|l|l|l|l|l|l|l|}
\hline
\rowcolor{Gray}
Tick & Neuron(s) & Input axon(s)  & $v_j(t)$ & $l_j$ &$v_j^+$ & $s_j(t)$ \\ \hline
\multirow{3}{*}{1} & $init_{it}$, $init_v$ & $en$ & 1 & 0 & 1 & 1 \\                          \cline{2-7}
                   & $r_{0_{fb}}, r_{4_{fb}}, r_{5_{fb}}$ & $r_0, r_4, r_5$ & 1 & 0 & 1 & 1 \\ \cline{2-7}
                   & $r_{0_v}, r_{4_v}, r_{5_v}$ & $r_0, r_4, r_5$ & 1 & 0 & 1 & 1 \\ \hline
\multirow{2}{*}{2} & $r_{0_{fb}}, r_{4_{fb}}, r_{5_{fb}}$ & \underbar{$r_0$} \underbar{$r_4$},                               \underbar{$r_5$} & 1 & 0 & 1 & 1 \\ 
                    \cline{2-7}
                    & $r_{0_v}, r_{4_v}, r_{5_v}$ & \underbar{$r_0$}, \underbar{$r_4$}, \underbar{$r_5$},  & 1 & 0 & 1 & 1 \\                          \hline
                   
\end{tabular}
}
\end{table}

Following our example input of $10001100$, Table~\ref{tab:inputneurons} shows the execution for neurons in the \emph{Input Core} that receive spikes (input value of 1). In the first tick, we see the arrival of the input word on axons $r_0$,$r_4$, and $r_5$. These spikes trigger the feedback neurons ($r_{0_{fb}}$, $r_{4_{fb}}$, and $r_{5_{fb}}$) and the VNU neurons ($r_{0_{v}}$, $r_{4_{v}}$, and $r_{5_{v}}$) to spike in this tick. On the next tick and subsequent ticks, the feedback axons (\underbar{$r_0$}, \underbar{$r_4$}, and \underbar{$r_5$}) fire, creating a feedback loop by sending the input spikes back to the feedback neurons. They also connect to the VNU neurons, sending the data to the \emph{VNU Core} every tick. Connected to the enable axon $en$ are initialization neurons $init_{it}$ and $init_v$. These neurons signal the \emph{Iteration Counter Core} and \emph{VNU Core} respectively to initialize their execution. The reset axon $rst_{in}$ has a negative weight for all neurons that stop the feedback loop, except neurons $rst_{it}$ and $rst_v$, which forward the reset signal to the \emph{Iteration Counter Core} and \emph{VNU Core}. We send a signal on the reset axon to put the neurons back to their initial state before decoding another input word.

\subsection{VNU Core}\label{subsec:vnu}
\begin{figure}[t]
    \centering
    \includegraphics[width=0.85\linewidth]{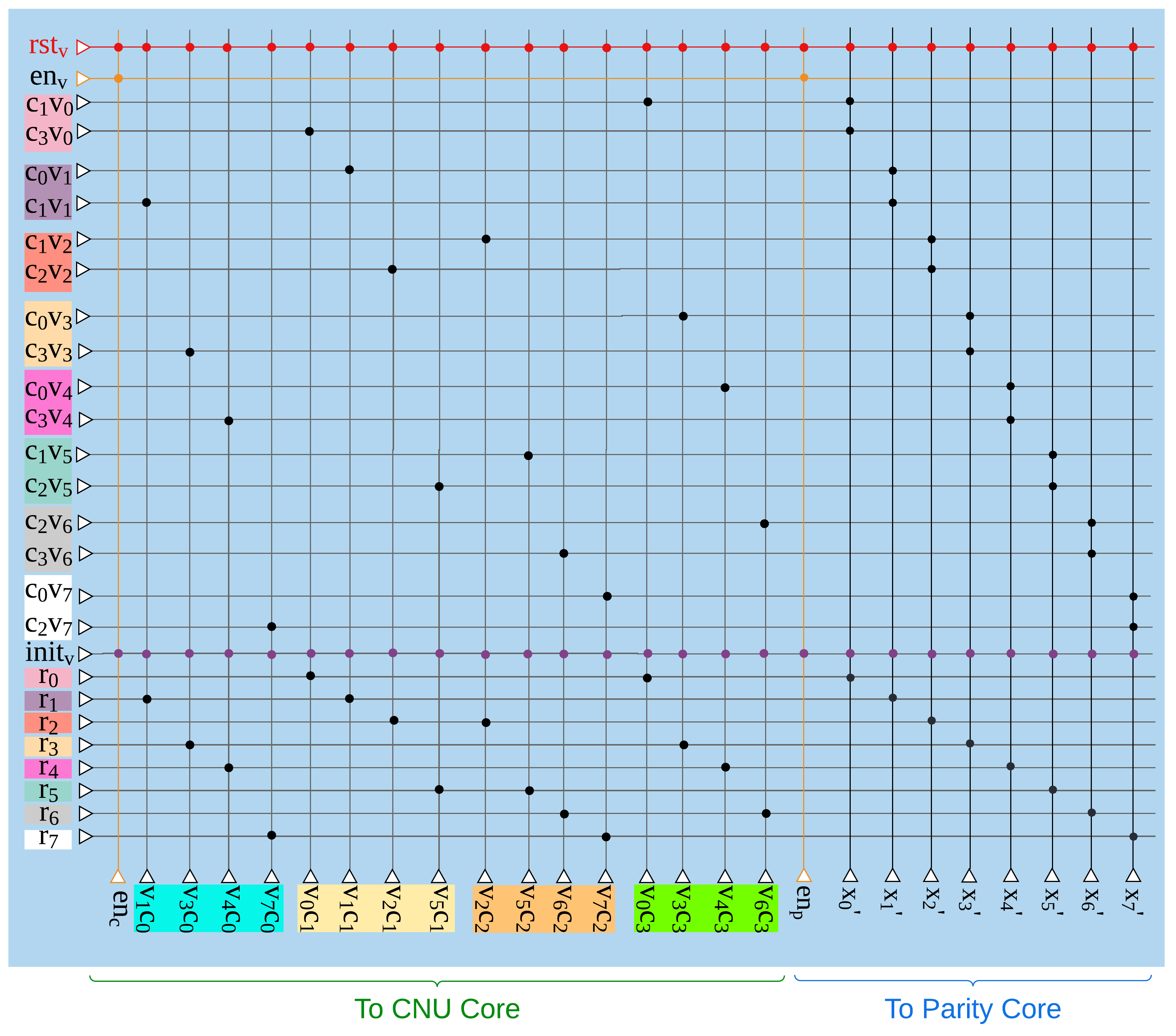}
    \caption{VNU Core.}
    \label{fig:vnucore}
    \vspace{-1mm}
\end{figure}

The \emph{VNU Core} implements 2 and 3-input majority voting function (Equation~\ref{eq:vnutocnu} and~\ref{eq:decide}) shown in steps 4 and 5 of the Algorithm~\ref{alg:gab}. The \emph{VNU Core} realizes a fully parallel implementation of eight VNUs. This core acts as a register in the first iteration corresponding to step 1 of Algorithm~\ref{alg:gab}, and during the subsequent iterations it implements 2-input majority voting function (Equation~\ref{eq:vnutocnu}) shown in step 4. Since each VNU generates three $v_nc_m$ output signals, a total of 24 such $v_nc_m$ signals are forwarded to the \emph{CNU Core}. This core forwards a cross-core synchronization signal $en_v$ to the \emph{CNU Core} via neuron $en_c$ and the \emph{Parity Core} via neuron $en_p$. This core is illustrated in Figure~\ref{fig:vnucore} with 27 axons. These axons as per the figure from top to bottom are: reset axon $rst_v$, $en_v$ enable axon, 16 $c_mv_n$ axons that receive signals from the \emph{CNU Core}, along with an $init_v$ axon and 8 $r_0-r_7$ axons that receive signals from the \emph{Input Core}. The 16 $c_mv_n$ axons and 8 $r_0-r_7$ axons are categorized by 8 colors, where each color represents connections to one VNU. Therefore, axons of a single color are incoming signals from a single VNU. 
\begin{table}[t]
\vspace{-3mm}
\caption{Axon and Neuron configurations of \emph{VNU Core}.}
\vspace{-2mm}
\label{tab:vnu_axneuconfig}
\tabcolsep=0.18cm
\centering
\scalebox{\tabsize}{
\begin{tabular}{|l|l|l|l|l|l|l|}
\hline
\rowcolor{Gray}
\multicolumn{7}{|c|}{Axons} \\ \hline
\multicolumn{5}{|l}{Name} & \multicolumn{2}{|l|}{Type $\tau$} \\ \hline
\multicolumn{5}{|l}{$rst_v$} & \multicolumn{2}{|l|}{2}  \\ \hline
\multicolumn{5}{|l}{$en_v$ - $init_v$} & \multicolumn{2}{|l|}{0}  \\ \hline
\multicolumn{5}{|l}{$r_0-r_7$} & \multicolumn{2}{|l|}{1} \\ \hline
\rowcolor{Gray}
\multicolumn{7}{|c|}{Neurons} \\ \hline
Name & $v^+_j$ & $l_j$ & $r^+_j$ & Rst type & Op Type & $w_j$ \\ \hline
$en_{c}, en_p$ & 1 & 0 & 0 & Hard & 0 & $[1,2,-2,1]$\\ \hline
$v_1c_0 - v_6c_3$ & 1 & -1 & 0 & Hard & 0 & $[1,2,-2,1]$\\ \hline
$x'_0 - x'_7$ & 1 & -1 & 0 & Hard & 0 & $[1,1,-2,1]$\\ \hline
\end{tabular}
}
\end{table}

This core has 26 neurons, consisting of 2 enable $en_c$ and $en_p$ neurons that forward the enable signal on axon $en$ to the \emph{CNU Core} and the \emph{Parity Core}, 16 $v_nc_m$ neurons, all of which send signals to the \emph{CNU Core}, and 8 $x_0'-x_7'$ neurons that represent the output decision word. The $v_nc_m$ neurons are arranged from left to right as per their destination CNUs $c_0-c_3$ and are grouped by colors such that neurons of the same color represent VNU connections to a single CNU. Details on the neuron configurations are listed in Table~\ref{tab:vnu_axneuconfig}. The synaptic connections of the \emph{VNU Core} are arranged according to the Tanner graph shown in Figure~\ref{fig:hmat_tanner}. Without loss of generality, VNU $v_0$ in the Tanner graph is connected with CNUs $c_1$ and $c_3$ through $v_0c_1$ and $v_0c_3$ edges respectively. Correspondingly, the core-level design of VNU $v_0$ is represented by the pink colored axons in Figure~\ref{fig:vnucore}, and is split across 2 neurons, $v_0c_1$ and $v_0c_3$. Each of these neurons also has synaptic connections with $r_0$. Furthermore, $r_0$ and $c_3v_0$ are included in the synaptic connection of neuron $v_0c_1$ to calculate $v_0c_1$ signal.  

\begin{table}[t]
\caption{\emph{VNU Core} neuron activity, assuming input word $10001100$. }
\vspace{-2mm}
\label{tab:vnuneurons}
\tabcolsep=0.18cm
\centering
\scalebox{\tabsize}{
\begin{tabular}{|l|l|l|l|l|l|l|}
\hline
\rowcolor{Gray}
Tick & Neuron(s) & Input axon(s)  & $v_j(t)$ & $l_j$ & $v_j^+$ & $s_j(t)$ \\ \hline
\multirow{4}{*}{2} & $en_{c}, en_p$ & $init_v$ & 1 & 0 & 1 & 1 \\                          \cline{2-7}
                   & $v_0c_1, v_0c_3, x_0'$ & $init_v, r_0$ & 3 & -1 & 1 & 1 \\ \cline{2-7}
                   & $v_4c_0, v_4c_3, x_4'$ & $init_v, r_4$ & 3 & -1 & 1 & 1 \\ \cline{2-7}
                   & $v_5c_1, v_5c_2, x_5'$ & $init_v, r_5$ & 3 & -1 & 1 & 1 \\ \hline
\multirow{9}{*}{4} & $en_{c}, en_p$ & $en_v$ & 1 & 0 & 1 & 1 \\                          \cline{2-7}
                   & $x'_0$ & $r_0, c_1v_0, c_3v_0$ & 3 & -1 & 1
                            & 1 \\                          \cline{2-7}
                   & $x'_1$ & $c_0v_1$ & 1 & -1 & 1
                            & 0 \\                          \cline{2-7}
                    & $x'_2$ & $c_2v_2$ & 1 & -1 & 1
                            & 0 \\                          \cline{2-7}
                    & $x'_3$ & $c_0v_3$ & 1 & -1 & 1
                            & 0 \\                          \cline{2-7}
                    & $x'_4$ & $r_4, c_3v_4$ & 2 & -1 & 1
                            & 1 \\                          \cline{2-7}
                    & $x'_5$ & $r_5, c_1v_5$ & 2 & -1 & 1
                            & 1 \\                          \cline{2-7}
                    & $x'_6$ & $c_2v_6$ & 1 & -1 & 1
                            & 0 \\                          \cline{2-7}
                    & $x'_7$ & $c_0v_7, c_2v_7$ & 2 & -1 & 1
                            & 1 \\                          \hline
                   
\end{tabular}
}
 \vspace{-2mm}
\end{table}

The tick-by-tick neuron execution of this core following our example input is shown in Table~\ref{tab:vnuneurons}. All neurons receive a spike via the $init_v$ axon on tick 2, so we only list the neurons that spike in this tick for brevity. In tick 2, the input word is forwarded to the \emph{Parity Core} by neurons ${x'_0, x'_4, x'_5}$ before any GaB correction begins to check if the input word needs correction. In parallel, the first iteration of VNU suggestions are sent to the \emph{CNU Core} via neurons ${v_0c_1, v_0c_3, v_4c_0, v_4c_3, v_5c_1, v_5c_2}$. During tick 3 the \emph{VNU Core} is waiting for the suggestions from the \emph{CNU Core}, so there is no relevant activity on the \emph{VNU Core} for this tick, however spikes still arrive from the \emph{Input Core} with the original input word. Therefore, the neuron leak values and enable neurons, $en_c$ and $en_p$, both ensure that any axon signals during this tick do not impact the execution. During tick 4, the CNU suggestions are fed into the relevant VNU axons for computing the VNU suggestions of the next GaB iteration, and computing the decision word on neurons $x'_0 - x'_7$. In Table~\ref{tab:vnuneurons}, we only show the values for the decision neurons as to highlight a single GaB iteration. Subsequent GaB iterations occur as described in tick 2, with the difference being that there is no $init$ signal, so spikes only occur based on CNU suggestions and initial $r_i$ values. The decision word after the first iteration is on neurons $x'_0 - x'_7$ as shown in the table, which is the corrected codeword $10001101$ that will be passed to the \emph{Parity Core} and subsequent cores for verification.

\subsection{CNU Core}
\begin{figure}
    \centering
    \begin{subfigure}[c]{0.53\linewidth}
        \includegraphics[width=\linewidth, trim={0cm 0.5cm 0cm 0.5cm}, clip=true]{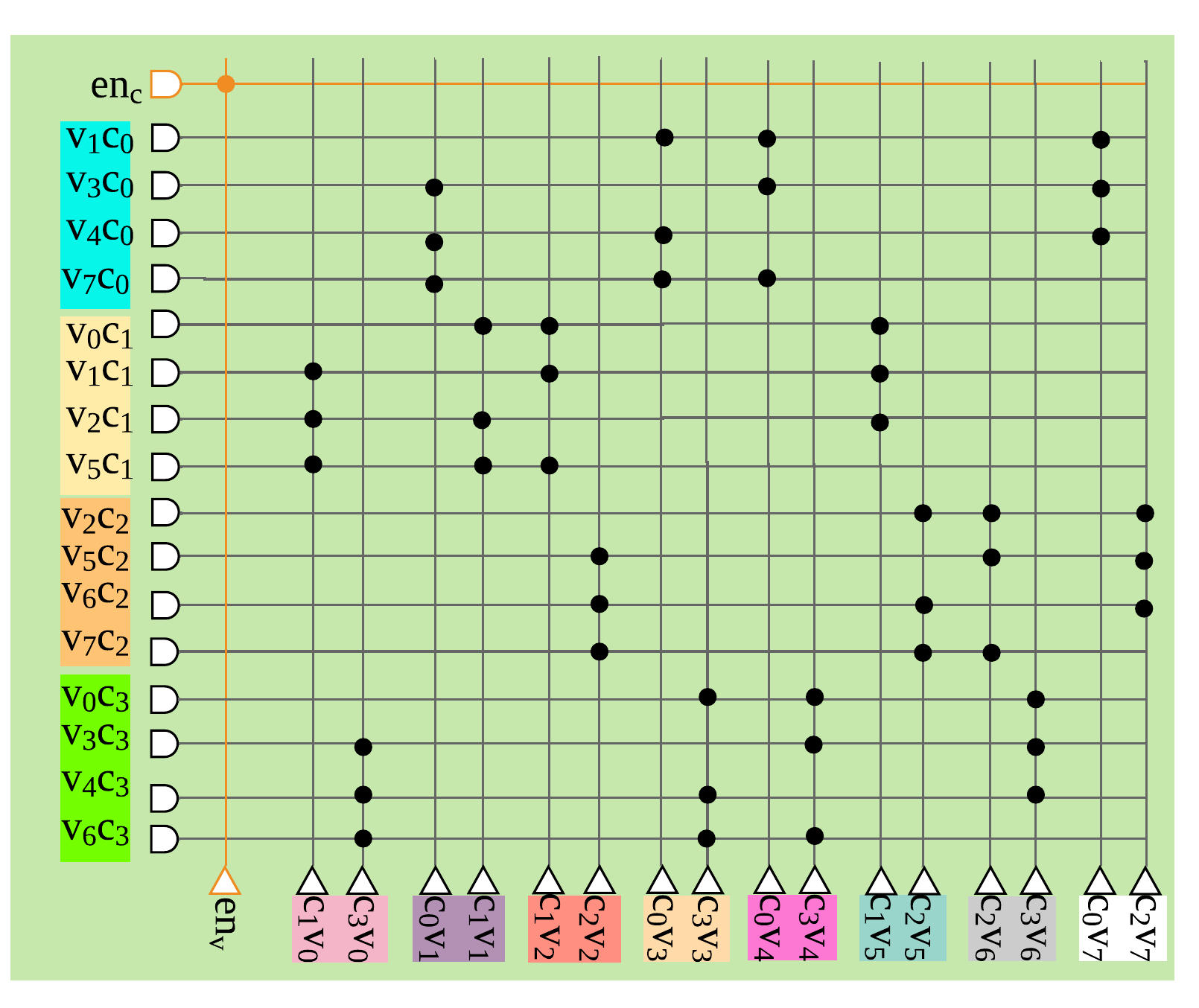}
        \caption{}
        \label{fig:cnucore}
    \end{subfigure}
    \begin{subfigure}[c]{0.2\linewidth}
        \includegraphics[width=\linewidth]{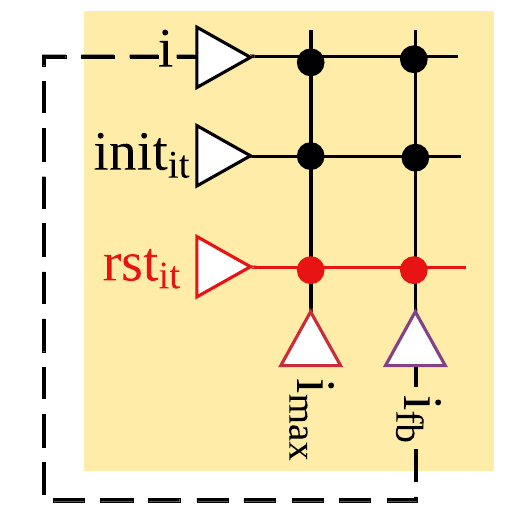}
        \caption{}
        \label{fig:iterationcountcore}
    \end{subfigure}
    \caption{(a) CNU Core, (b) Iteration Counter Core.}
\end{figure}

The \emph{CNU Core} realizes a fully parallel implementation of four CNUs as illustrated in Figure~\ref{fig:cnucore}. Each CNU calculates suggestions for its four connected VNUs, as per the Tanner graph in Figure~\ref{fig:hmat_tanner}, through a 3-input XOR function as shown in Equation~\ref{eq:cnu} corresponding to step 3 of Algorithm~\ref{alg:gab}. The \emph{CNU Core} sends its suggestions to the \emph{VNU Core} for iterative execution of step 4 and step 5 of the Algorithm~\ref{alg:gab}. The 16 $v_nc_m$ signals received in this core from the \emph{VNU Core} are represented as the axons, color coded and grouped by CNU. Including the $en_c$ synchronization signal, the \emph{CNU Core} receives a total of 17 signals from the \emph{VNU Core}. All of the neurons of this core have a threshold of 1 and reset value of 0 as shown in Table~\ref{tab:cnu_axneuconfig}. Table also highlights the neurons implementing the XOR operation with the Op Type field. The \emph{CNU Core} sends the 17 neuron outputs back to the \emph{VNU Core} to execute step 5 of Algorithm~\ref{alg:gab}. The execution of the \emph{CNU Core} following the example input word $10001100$ is shown in Table~\ref{tab:cnuneurons}. Suggestions from the \emph{VNU Core} during tick 2 arrive on the axons of this core at tick 3. The 3-input xor operation is calculated across all the connected axons for each of the $c_mv_n$ neurons in a single tick due to the modified neuron block. Neurons whose xor result yield a value of 1 fire, sending CNU suggestion back to the respective VNU axons for computing the decision value on tick 4 as described in Section~\ref{subsec:vnu}.

\begin{table}[t]
\caption{Axon and Neuron configurations of \emph{CNU Core}.}\vspace{-2mm}
\label{tab:cnu_axneuconfig}
\tabcolsep=0.18cm
\centering
\scalebox{\tabsize}{
\begin{tabular}{|l|l|l|l|l|l|l|}
\hline
\rowcolor{Gray}
\multicolumn{7}{|c|}{Axons} \\ \hline
\multicolumn{5}{|l}{Name} & \multicolumn{2}{|l|}{Type $\tau$} \\ \hline
\multicolumn{5}{|l}{$en_c - v_6c_3$} & \multicolumn{2}{|l|}{0}  \\ \hline
\rowcolor{Gray}
\multicolumn{7}{|c|}{Neurons} \\ \hline
Name & $v^+_j$ & $l_j$ & $r^+_j$ & Rst type & Op Type & $w_j$ \\ \hline
$en_v$ & 1 & 0 & 0 & Hard & 0 & $[1,1,1,1]$\\ \hline
$v_1c_0 - v_6c_3$ & 1 & 0 & 0 & Hard & 1 & $[1,1,1,1]$\\ \hline
\end{tabular}
}
\vspace{-1mm}
\end{table}

\begin{table}[t]
\caption{\emph{CNU Core} neuron activity, following input word $10001100$. The only relevant execution for the first GaB iteration is during tick 3 as shown.}
\vspace{-2mm}
\label{tab:cnuneurons}
\tabcolsep=0.18cm
\centering
\scalebox{\tabsize}{
\begin{tabular}{|l|l|l|l|l|l|l|}
\hline
\rowcolor{Gray}
Tick & Neuron(s) & Input axon(s)  & $v_j(t)$ & $l_j$ &$v_j^+$ & $s_j(t)$ \\ \hline
\multirow{10}{*}{3} & $en_v$ & $en_c$ & 1 & 0 & 1 & 1 \\                          \cline{2-7}
                    & $c_1v_0$ & $v_5c_1$ & 1 & 0 & 1 & 1 \\ \cline{2-7}
                    & $c_3v_0$ & $v_4c_3$ & 1 & 0 & 1 & 1 \\ \cline{2-7}
                    & $c_0v_1, c_0v_3, c_0v_7$ & $v_4c_0$ & 1 & 0 & 1 & 1 \\ \cline{2-7}
                    & $c_{1}v_{1}, c_{1}v_{2}$ & $v_{0}c_{1}, v_{5}c_{1}$ & 0 & 0 & 1 & 0 \\ \cline{2-7}
                    & $c_2v_2, c_2v_6, c_2v_7$ & $v_5c_2$ & 1 & 0 & 1 & 1 \\ \cline{2-7}
                    & $c_3v_3, c_3v_6$ & $v_0c_3, v_4c_3$ & 0 & 0 & 1 & 0 \\ \cline{2-7}
                    & $c_3v_4$ & $v_0c_3$ & 1 & 0 & 1 & 1 \\ \cline{2-7}
                    & $c_1v_5$ & $v_0c_1$ & 1 & 0 & 1 & 1 \\ \cline{2-7}
                    & $c_0v_4, c_2v_5$ & $-$ & 0 & 0 & 1 & 0 \\ \hline
                
\end{tabular}
}
\end{table}

\subsection{Iteration Counter Core}\label{subsec:iteration_core}
The \emph{Iteration Counter Core}, shown in Figure~\ref{fig:iterationcountcore}, executes step 2 of Algorithm~\ref{alg:gab}, by keeping track of the number of GaB iterations, and producing a spike if the iteration count reaches $maxIter$. Axon $i$ receives its signal as part of a feedback loop and the $i_0$ axon receives its signal from the \emph{Input Core} to start the iteration count. The $rst_{it}$ axon is used to reset the core, with negative weights selected on the neurons large enough to set their potential back to 0 without spiking as shown in Table~\ref{tab:itercount_axneuconfig}. The feedback neuron $i_{fb}$ serves as a register, incrementing the potential of $i_{max}$ by one every tick. The neuron $i_{max}$ functions as a counter and a comparator, with a positive threshold set to the total number of ticks required to achieve the maximum GaB iteration count $maxIter$. The expression to determine the positive threshold $v_j^+$ for neuron $i_{max}$ is shown in Equation~\ref{eq:iter_calc}.

\footnotesize
\setlength{\textfloatsep}{0pt}
\begin{equation}\label{eq:iter_calc}
((maxIter-1) \times 2)+4
\end{equation} 
\normalsize

\begin{table}[b]
\vspace{2mm}
\caption{Axon and Neuron configurations of \emph{Iteration Counter Core}.}
\vspace{-2mm}
\label{tab:itercount_axneuconfig}
\tabcolsep=0.18cm
\centering
\scalebox{\tabsize}{
\begin{tabular}{|l|l|l|l|l|l|l|}
\hline
\rowcolor{Gray}
\multicolumn{7}{|c|}{Axons} \\ \hline
\multicolumn{5}{|l}{Name} & \multicolumn{2}{|l|}{Type $\tau$} \\ \hline
\multicolumn{5}{|l}{$i, init_{it}$} & \multicolumn{2}{|l|}{0}  \\ \hline
\multicolumn{5}{|l}{$rst_{it}$} & \multicolumn{2}{|l|}{1} \\ \hline
\rowcolor{Gray}
\multicolumn{7}{|c|}{Neurons} \\ \hline
Name & $v^+_j$ & $l_j$ & $r^+_j$ & Rst type & Op Type & $w_j$ \\ \hline
$i_{max}$ & 202 & 0 & 0 & Hard & 0 & $[1,-202,1,1]$\\ \hline
$i_{fb}$ & 1 & 0 & 0 & Hard & 0 & $[1,-1,1,1]$\\ \hline

\end{tabular}
}
\end{table}

Decision process during each iteration involves VNU suggestions followed by CNU suggestions. Therefore, a total of 2 ticks are required per iteration, resulting in the factor of 2 in Equation~\ref{eq:iter_calc}. After the last GaB iteration, there is a 2 core offset until the decision reaches the OR core, which is when we want the spike to arrive from the \emph{Iteration Counter Core}. Therefore, for $maxIter-1$ iterations, only 2 ticks are required, but the last GaB iteration requires 2 ticks plus a 2 tick offset. 

Following our running example in Table~\ref{tab:itercountneurons}, the core receives its initial spike to start counting from the \emph{Input Core} on axon $init_{it}$ at tick 2. This axon signal starts the feedback loop with neuron $i_{fb}$ and begins the count for $i_{max}$. In subsequent ticks, neuron $i_{fb}$ will continue to spike and send its signal onto axon $i$ where the potential in neuron $i_{max}$ will increment until it reaches the positive threshold set. After this threshold is reached, the neuron will spike to the \emph{OR Core} indicating that the maximum GaB iteration count was reached. The count will then start over unless a signal is sent to $rst_{it}$.

\begin{table}[t]
\caption{\emph{Iteration Counter Core} neuron activity for 100 GaB iterations.}
\vspace{-2mm}
\label{tab:itercountneurons}
\tabcolsep=0.18cm
\centering
\scalebox{\tabsize}{
\begin{tabular}{|l|l|l|l|l|l|l|}
\hline
\rowcolor{Gray}
Tick & Neuron(s) & Input axon(s)  & $v_j(t)$ & $l_j$ &$v_j^+$ & $s_j(t)$ \\ \hline
\multirow{2}{*}{2} & $i_{fb}$ & $init_{it}$ & 1 & 0 & 1 & 1 \\                          \cline{2-7}
                   & $i_{max}$ & $init_{it}$ & 1 & 0 & 202 & 0   \\ \hline
\multirow{2}{*}{3} & $i_{fb}$ & $i$ & 1 & 0 & 1 & 1 \\                          \cline{2-7}
                   & $i_{max}$ & $i$ & 2 & 0 & 202 & 0   \\ \hline
\multirow{2}{*}{4} & $i_{fb}$ & $i$ & 1 & 0 & 1 & 1 \\                          \cline{2-7}
                   & $i_{max}$ & $i$ & 3 & 0 & 202 & 0   \\ \hline
\multicolumn{7}{|c|}{...}\\ \hline

\multirow{2}{*}{202} & $i_{fb}$ & $i$ & 1 & 0 & 1 & 1 \\                          \cline{2-7}
                   & $i_{max}$ & $i$ & 201 & 0 & 202 & 0   \\ \hline
\multirow{2}{*}{203} & $i_{fb}$ & $i$ & 1 & 0 & 1 & 1 \\                          \cline{2-7}
                   & $i_{max}$ & $i$ & 202 & 0 & 202 & 1  \\ \hline

\end{tabular}
}
\vspace{-2mm}
\end{table}

\begin{table}[t]
\caption{Axon and Neuron configurations of \emph{Parity Core}.}
\vspace{-2mm}
\label{tab:parity_axneuconfig}
\tabcolsep=0.18cm
\centering
\scalebox{\tabsize}{
\begin{tabular}{|l|l|l|l|l|l|l|}
\hline
\rowcolor{Gray}
\multicolumn{7}{|c|}{Axons} \\ \hline
\multicolumn{5}{|l}{Name} & \multicolumn{2}{|l|}{Type $\tau$} \\ \hline
\multicolumn{5}{|l}{$en_p - x'_7$} & \multicolumn{2}{|l|}{0}  \\ \hline
\rowcolor{Gray}
\multicolumn{7}{|c|}{Neurons} \\ \hline
Name & $v^+_j$ & $l_j$ & $r^+_j$ & Rst type & Op Type & $w_j$ \\ \hline
$en_s, x'_0 - x'_7$ & 1 & 0 & 0 & Hard & 0 & $[1,1,1,1]$\\ \hline
$s_0 - s_3$ & 1 & 0 & 0 & Hard & 1 & $[1,1,1,1]$\\ \hline
\end{tabular}
}
\end{table}

\begin{table}[t]
\vspace{-2mm}
\caption{\emph{Parity Core} neuron activity, assuming input word $10001100$. GaB iterations 0 (tick 3) and 1 (tick 5) are shown for comparison.}
\vspace{-2mm}
\label{tab:parityneurons}
\tabcolsep=0.18cm
\centering
\scalebox{\tabsize}{
\begin{tabular}{|l|l|l|l|l|l|l|}
\hline
\rowcolor{Gray}
Tick & Neuron(s) & Input axon(s)  & $v_j(t)$ & $l_j$ & $v_j^+$ & $s_j(t)$ \\ \hline
\multirow{8}{*}{3} & $en_s$ & $en_p$ & 1 & 0 & 1 & 1 \\                          \cline{2-7}
                   & $s_0$ & $x'_4$ & 1 & 0 & 1 & 1 \\ \cline{2-7}
                   & $s_1$ & $x'_0, x'_5$ & 0 & 0 & 1 & 0 \\ \cline{2-7}
                   & $s_2$ & $x'_5$ & 1 & 0 & 1 & 1 \\ \cline{2-7}
                   & $s_3$ & $x'_0, x'_4$ & 0 & 0 & 1 & 0 \\ \cline{2-7}
                   & $x'_0$ & $x'_0$ & 1 & 0 & 1 & 1 \\ \cline{2-7}
                   & $x'_4$ & $x'_4$ & 1 & 0 & 1 & 1 \\ \cline{2-7}
                   & $x'_5$ & $x'_5$ & 1 & 0 & 1 & 1 \\ \hline
\multirow{9}{*}{5} & $en_s$ & $en_p$ & 1 & 0 & 1 & 1 \\                          \cline{2-7}
                   & $s_0$ & $x'_4, x'_7$ & 0 & 0 & 1 & 0 \\ \cline{2-7}
                   & $s_1$ & $x'_0, x'_5$ & 0 & 0 & 1 & 0 \\ \cline{2-7}
                   & $s_2$ & $x'_5, x'_7$ & 0 & 0 & 1 & 0 \\ \cline{2-7}
                   & $s_3$ & $x'_0, x'_4$ & 0 & 0 & 1 & 0 \\ \cline{2-7}
                   & $x'_0$ & $x'_0$ & 1 & 0 & 1 & 1 \\ \cline{2-7}
                   & $x'_4$ & $x'_4$ & 1 & 0 & 1 & 1 \\ \cline{2-7}
                   & $x'_5$ & $x'_5$ & 1 & 0 & 1 & 1 \\ \cline{2-7}
                   & $x'_7$ & $x'_7$ & 1 & 0 & 1 & 1 \\ \hline
                   
\end{tabular}
}
\end{table}

\begin{table}[t]
\vspace{-1mm}
\caption{Axon and Neuron configurations of \emph{Syndrome Core}.}
\vspace{-2mm}
\label{tab:syndrome_axneuconfig}
\tabcolsep=0.18cm
\centering
\scalebox{\tabsize}{
\begin{tabular}{|l|l|l|l|l|l|l|}
\hline
\rowcolor{Gray}
\multicolumn{7}{|c|}{Axons} \\ \hline
\multicolumn{5}{|l}{Name} & \multicolumn{2}{|l|}{Type $\tau$} \\ \hline
\multicolumn{5}{|l}{$en_s$} & \multicolumn{2}{|l|}{0}  \\ \hline
\multicolumn{5}{|l}{$s_0-s_3$} & \multicolumn{2}{|l|}{1}  \\ \hline
\rowcolor{Gray}
\multicolumn{7}{|c|}{Neurons} \\ \hline
Name & $v^+_j$ & $l_j$ & $r^+_j$ & Rst type & Op Type & $w_j$ \\ \hline
$zero$ & 1 & 0 & 0 & Hard & 0 & $[1,-1,1,1]$\\ \hline
\end{tabular}
}
\end{table}

\subsection{Parity and Syndrome Core}

\begin{figure}[b]
    \centering
    \begin{subfigure}[c]{0.6\linewidth}
        \centering
        \includegraphics[width=0.9\linewidth]{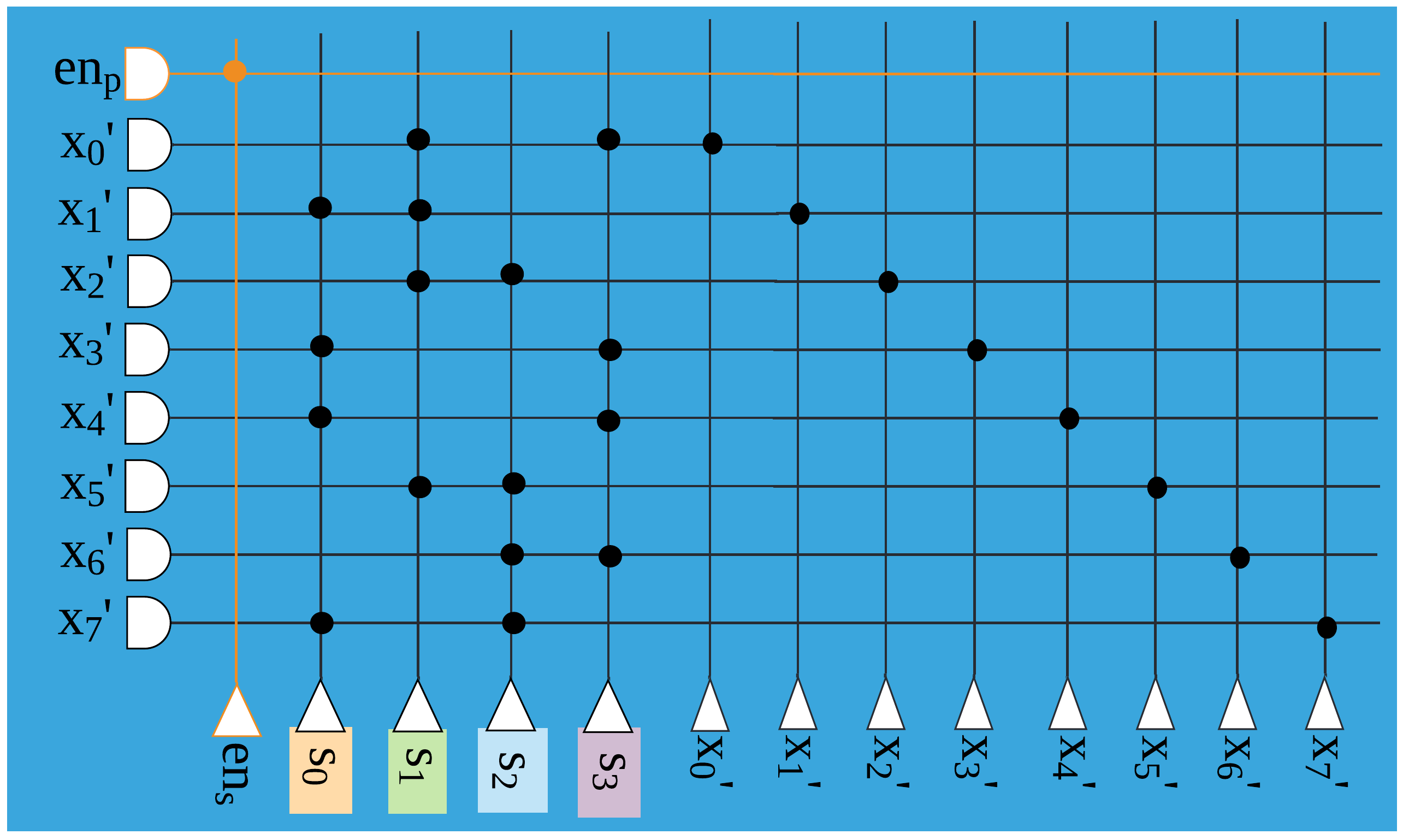}
        \caption{}
        \label{fig:paritycore}
    \end{subfigure}
    \begin{subfigure}[c]{0.21\linewidth}
        \centering
        \includegraphics[width=0.9\linewidth]{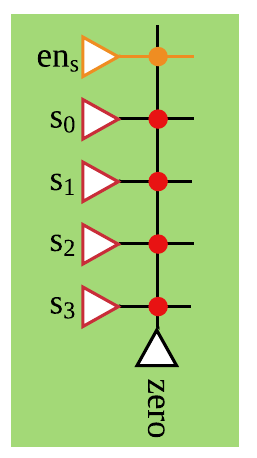}
        \caption{}
        \label{fig:syndromecore}
    \end{subfigure}
    \caption{(a) Parity Core, (b) Syndrome Core.}
\end{figure}

Based on step 6 (Equation~\ref{eq:syndrome}) of Algorithm~\ref{alg:gab}, the \emph{Parity Core} executes the $mod$ $2$ VMM and the \emph{Syndrome Core} checks if the generated product is an all-zero vector by implementing a 4-input NOR operation. The \emph{Parity Core} shown in Figure~\ref{fig:paritycore} has 9 axons, among which the 8 axons marked by $x^\prime_0-x^\prime_7$ are used to receive the decision word estimation from the \emph{VNU Core}, and the $en_p$ axon is used to receive spike for synchronization. This core has 13 neurons: the $en_s$ neuron, 4 $s_0-s_3$ neurons and 8 $x'_0-x'_7$ neurons. The $s_0-s_3$ neurons perform $mod$ $2$ addition operation by implementing the 4-input XOR function. The $x'_0-x'_7$ neurons forward the decision word to the \emph{Output Core}. All of the neurons have a threshold of 1 and reset to 0 after spiking, as shown in Table~\ref{tab:parity_axneuconfig}. The synaptic connections of this core reflect the $H$ matrix from Figure~\ref{fig:hmat_tanner}. Table~\ref{tab:parityneurons} shows execution of the \emph{Parity Core}. In tick 3, the received input results in non-zero $s_0$ and $s_2$, indicating incorrect estimation. In tick 5, $s_0-s_3$ are all-zero, meaning the input is correct codeword. There's a 2-tick spike delay at the neurons of this core, to wait for the \emph{Syndrome Core} operation to be complete.

The \emph{Syndrome Core} shown in Figure~\ref{fig:syndromecore} has one $en_s$ axon and four $s_0-s_3$ axons from the \emph{Parity Core}. This core implements a 4-input NOR operation among $s_0-s_3$, as well as an AND with the $en_s$ axon, such that the single neuron labeled $zero$ in the core only spikes when the $en_s$ axon receives a spike and the remaining axons receive 0. The axon and neuron configuration for this core is shown in Table~\ref{tab:syndrome_axneuconfig}. Since a negative neuron potential produces no spike, we assign negative weight values to the axons connected to the $s_0-s_3$ parity check neurons, such that one or more of them receiving an input signal is enough to counteract the positive signal from the $en_s$ axon. Table~\ref{tab:syndromeneurons} shows that, on tick 4, $s_0$ and $s_2$ axons receive spike, hence $zero$ neuron does not spike. On tick 6, $s_0-s_3$ being all-zero, a spike is produced.

\begin{table}[h]
\vspace{-2mm}
\caption{\emph{Syndrome Core} neuron activity, assuming input word $10001100$. GaB iterations 0 (tick 4) and 1 (tick 6) are shown for comparison.}
\vspace{-2mm}
\label{tab:syndromeneurons}
\tabcolsep=0.18cm
\centering
\scalebox{\tabsize}{
\begin{tabular}{|l|l|l|l|l|l|l|}
\hline
\rowcolor{Gray}
Tick & Neuron(s) & Input axon(s)  & $v_j(t)$ & $l_j$ & $v_j^+$ & $s_j(t)$ \\ \hline
\multirow{1}{*}{4} & $zero$ & $en_s, s_0, s_2$ & -1 & 0 & 1 & 0 \\                                    \hline
\multirow{1}{*}{6} & $zero$ & $en_s$ & 1 & 0 & 1 & 1 \\                          \hline
                   
\end{tabular}
}
\end{table}

\begin{figure}[b]
    \centering
    \begin{subfigure}[c]{0.3\linewidth}
        \includegraphics[width=\linewidth, trim={0cm 0cm 0cm 0.1cm},clip=true]{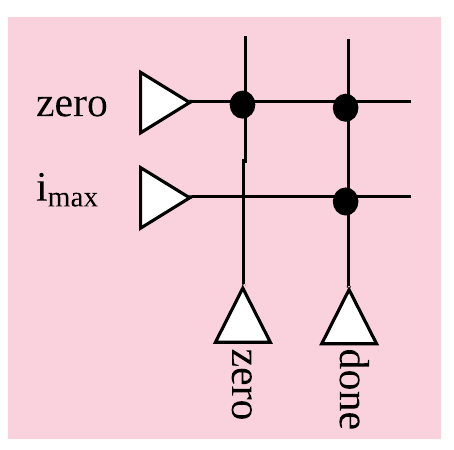}
        \caption{}
        \label{fig:orcore}
    \end{subfigure}
    \begin{subfigure}[c]{0.5\linewidth}
        \includegraphics[width=\linewidth, trim={0cm 0cm 0cm 0.1cm},clip=true]{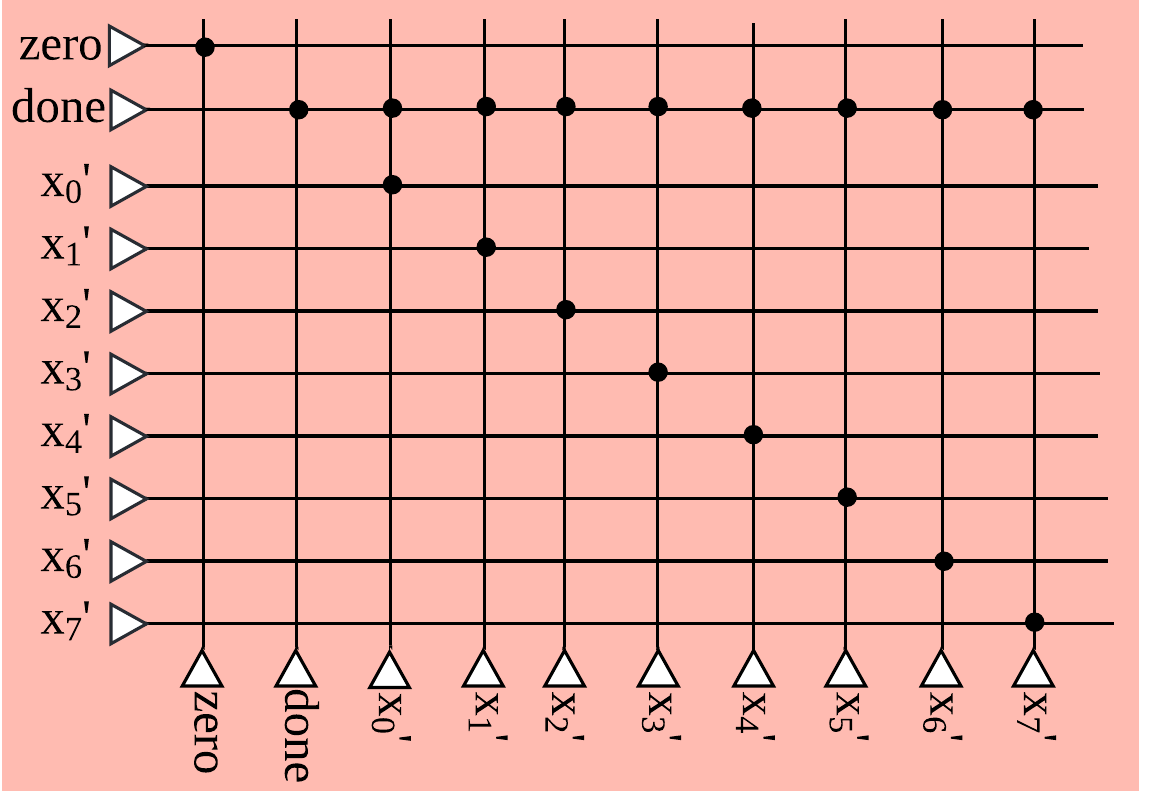}
        \caption{}
        \label{fig:outputcore}
    \end{subfigure}
    \caption{(a) OR Core, (b) Output Core.}
\end{figure}

\subsection{OR Core}
\begin{table}[t]
\caption{Axon and Neuron configurations of \emph{OR Core}.}
\vspace{-2mm}
\label{tab:or_axneuconfig}
\tabcolsep=0.18cm
\centering
\scalebox{\tabsize}{
\begin{tabular}{|l|l|l|l|l|l|l|}
\hline
\rowcolor{Gray}
\multicolumn{7}{|c|}{Axons} \\ \hline
\multicolumn{5}{|l}{Name} & \multicolumn{2}{|l|}{Type $\tau$} \\ \hline
\multicolumn{5}{|l}{$zero, i_{max}$} & \multicolumn{2}{|l|}{0}  \\ \hline
\rowcolor{Gray}
\multicolumn{7}{|c|}{Neurons} \\ \hline
Name & $v^+_j$ & $l_j$ & $r^+_j$ & Rst type & Op Type & $w_j$ \\ \hline
$zero, done$ & 1 & 0 & 0 & Hard & 0 & $[1,1,1,1]$\\ \hline
\end{tabular}
}
\end{table}

The \emph{OR Core} presented in Figure~\ref{fig:orcore} implements a logical OR operation between two received signals from the \emph{Iteration Counter Core} and \emph{Syndrome Core}. This operation checks if the maximum iteration count has been reached, or the input word has been corrected. The result of the OR operation indicated by the $done$ neuron is sent to the \emph{Output Core}. Additionally, the $zero$ neuron forwards the value from the $zero$ axon arriving from the \emph{Syndrome Core} to indicate whether the presented output is a valid codeword or not. The axon and neuron configurations are listed in Table~\ref{tab:or_axneuconfig} along with the tick-level execution in Table~\ref{tab:orneurons}.

\begin{table}[t]
\vspace{-2mm}
\caption{\emph{OR Core} neuron activity, assuming input word $10001100$. GaB iterations 0 (tick 5) and 1 (tick 7) are shown for comparison.}
\vspace{-2mm}
\label{tab:orneurons}
\tabcolsep=0.18cm
\centering
\scalebox{\tabsize}{
\begin{tabular}{|l|l|l|l|l|l|l|}
\hline
\rowcolor{Gray}
Tick & Neuron(s) & Input axon(s)  & $v_j(t)$ & $l_j$ & $v_j^+$ & $s_j(t)$ \\ \hline
\multirow{1}{*}{5}
                   & $zero, done$ & - & 0 & 0 & 1 & 0 \\ \hline
\multirow{1}{*}{7} & $zero, done$ & $zero$ & 1 & 0 & 1 & 1 \\ \hline
\end{tabular}
}
\end{table}

\subsection{Output Core}

The \emph{Output Core} (Figure~\ref{fig:outputcore}) is responsible for sending the decoded codeword to the off-chip host upon completion of the decoding process, along with two flags indicating the status of the result. There are 10 axons, among which 8 axons $x^\prime_0 - x^\prime_7$ are used to receive estimated codeword from the \emph{Parity Core}. This core has 8 neurons marked as $x^\prime_0 - x^\prime_7$, that send the estimated word to the off-chip host via an AND operation between the $x^\prime_0 - x^\prime_7$ input and the $done$ signal. The $zero$ neuron forwards the result from the $zero$ axon, indicating the correctness of the output word.

\section{Experimental Setup}
\label{sec:exp_setup}
In this study for our experimental evaluations, we utilize RANC~\cite{ranc}, a Reconfigurable Architecture for Neuromorphic Computing ecosystem through software based simulation and FPGA based emulation. RANC is an open-source, configurable emulation environment that allows for rapid prototyping of novel neuromorphic architectures for researchers to explore neuromorphic designs before committing to silicon, and without imposing the limitations and expense of pre-fabricated ASIC chips. RANC has been verified to be cycle-by-cycle accurate when emulating the TrueNorth chip\cite{Akopyan2015TrueNorth}, ensuring that the spiking functionality of the neuromorphic cores in the emulation environment match what would be expected on neuromorphic hardware. RANC supports the key operations of neuromorphic architectures and is highly parameterized with configurable components allowing application engineers and hardware architects to experiment with application mapping and hardware tuning concurrently. 

One of the key features that differentiates RANC from other neuromorphic simulation-oriented works is by providing an end-to-end ecosystem that tightly couples software simulation and hardware emulation. As such, we can combine simulation and FPGA based analysis to progressively move from testing new ideas in the software environment to implementing equivalent changes in hardware. This allows rapid validation of the application through software and enables evaluating further impacts on performance metrics such as power and resource utilization through hardware. Neuromorphic simulation platforms such as PyCARL~\cite{balaji2020pycarl}, NeMo~\cite{NeMo2018}, and SpykeTorch~\cite{mozafari2019spyketorch} are suitable for simulating the GaB application, however those platforms do not offer FPGA-based hardware emulation.

We select a dataset of 288 words, based on the 32 words supported by our H-matrix described in Section~\ref{subsec:8-bit_decoder} plus 1-bit noise combinations of those words. The H-Matrix defines the implementation of the decoder design for a neuromorphic platform as presented in Section~\ref{sec:mapping}, as well as the serial Python GaB implementation used for verification. Once the neuromorphic mapping is complete, the configuration files are passed to the RANC environment to execute. Finally, the output spikes are collected and compared with the expected output from the serial implementation. We found when implementing our GaB design that there was a direct input-output match between the serial implementation and the software/hardware environments over our dataset. Additionally, there was a tick-by-tick match between both the software and hardware based executions of the algorithm. The following subsections will describe in more detail the execution of the software simulation environment and the hardware emulation environment given the configuration parameters.

\subsection{Software Simulation Environment}
\label{subsec:software_simulation}
The RANC software simulator is a tick-accurate simulation of the RANC architecture. There are two configuration files that we provide to the simulator based on the decoder design presented in Section~\ref{sec:mapping}. The first file is a configuration file that specifies global parameters for the simulation such as the dimensions of the RANC NoC or flags for enabling or disabling debug logging. The second file specifies the remainder of the configuration required for proper simulator execution. This includes the input spikes that will be sent into the RANC NoC, representative of our input words, as well as the threshold, connection, and reset mode parameters for each neuron. We run the RANC simulator on an Intel Core i7-8700 3.20GHz CPU. After launching the simulator, a trace is collected containing all spikes sent to the \emph{Output Core}, which can then be analyzed for tick-count and functional verification.

\subsection{FPGA Emulation Environment}\label{subsec:fpga_emulation}

To utilize the FPGA emulation environment, RANC provides an FPGA IP wizard that allows for specifying a set of memory files to configure each of the cores present in a given design. After this, the design is synthesized that contains all information about the connections between neurons, as well as their threshold and spiking behaviors. The IP core can then be deployed into a broader design, utilizing AXI4 communication to route spikes and collect results from the \emph{Output Core}. RANC models the components of a neuromorphic architecture independently, allowing for easy modification of internal component behavior. We leverage this to make our proposed modification to the neuron block, leaving the remaining neuromorphic component functionalities unchanged. Output from the FPGA emulation can be directly compared tick-by-tick against the output from the simulation environment and the serial GaB implementation. The FPGA emulation environment also allows for gathering of performance metrics such as power, energy, and resource utilization. We primarily use these estimates as a point of comparison for our proposed XOR-integrated design, as we know that the actual estimated values are not a reflection of the expected values for deployment on real neuromorphic hardware. We collect resource utilization percentage from the LUTs, LUTRAMs, Flip-Flops, and BRAMs targeting the Zynq UltraScale+ MPSoC ZCU102. We also perform power estimation with the Vivado Power Analysis tool by providing switching activity files (SAIF files) gathered from running a subset of the dataset through the baseline and XOR emulated designs. We chose to perform this analysis on a subset of our overall dataset due to the prohibitive timescale of gathering switching activity for every input in our emulation environment. We considered two factors when determining the values of this subset: the convergence of the input and the binary makeup of the input. Both of these factors would have an impact on the results produced, so our general approach was to have an even distribution of both converging and non-converging inputs, as well as an even distribution of 1s and 0s in each input word. The size of the subset was determined to be the smallest number of pairs of converging and non-converging input words that produced representative power and resource utilization results. A subset is considered representative in our analysis when increasing the size of the subset does not include noticeable variations to the results, which we found to occur six data points. We measure total power to evaluate the impact of the XOR modification as we scale the design to support larger datasets, which require a larger neuromorphic grid of cores to compute. We further measure the average power per core to analyze the impact the XOR modification has on individual cores as we scale the design. Using the power estimates, the tick count, and tick frequency, we also calculate energy estimates for the deployment of the designs onto the ZCU102. To evaluate the same impacts from the power analysis with regards to energy, we show both total energy and average energy per core for each design at different grid dimensions. Additionally, we define two parameters (the total number of words to decode, and number of GaB iterations per word) that are innate to scaling towards larger datasets and have an impact on the execution time of the decoder. We run a sweeping experiment to evaluate how the energy estimates scale with the increased execution time defined by those parameters for realistically larger datasets.

\section{Results}
\label{sec:results}
We analyze our GaB design for two neuromorphic implementations, the first is using the baseline LIF neuron block presented in Section~\ref{sec:neuromorphic_model} with Figure~\ref{fig:neuron_activecomp}. The second is using the XOR modified neuron block shown in Figure~\ref{fig:neuronblock_xor} following the mapping approach presented in Section~\ref{sec:mapping}. We discuss the difference between the designs in terms of execution time (tick count), resource utilization, power estimation, and energy estimation on the ZCU102 evaluation board.

\subsection{Execution Time}

We determine execution time in the simulation environment in terms of simulation steps or ticks required to complete the decoding process for all inputs. The baseline and XOR-integrated designs require 87,556 ticks and 59,042 ticks respectively to process the 288 word dataset sequentially, resulting in a total tick reduction of 32.57\%. To determine the tick frequency on the emulation environment, we first have to allow for each synapse calculation to be executed on the core’s neuron block. For the baseline 256x256 axon neuron dimensions on a core, there are 65,536 synaptic calculations to complete, each of which requires a single clock cycle. Running at a board frequency of 100MHz, all of the calculations for a tick of this core size can be completed at a frequency of 1.5kHz. We chose 100MHz frequency in order to achieve a reasonable simulation time, as well as maintain a similar tick frequency to that of TrueNorth\cite{Akopyan2015TrueNorth} In emulation, the execution times are 39.36 seconds and 58.37 seconds, while in software-based simulation, execution time is much longer taking around 53 minutes and 80 minutes for the XOR-integrated and baseline designs respectively. The FPGA-based emulation reduces the timescale of the simulations by up to a factor of 82x. In order to see the scalability of both designs beyond our dataset and configuration, we determined a general equation to calculate the total tick requirement for the baseline design (Equation~\ref{eq:baselineTick}) and XOR-integrated design (Equation~\ref{eq:XORTick}) based on the number of GaB iterations per word ($maxIter$) and the number of words in the dataset ($w_c$) for each implementation.

\vspace{-0.4cm}
\footnotesize
\begin{equation}\label{eq:baselineTick}
    t_{b} = w_c*(3*maxIter+4) + 4
\end{equation}
\vspace{-0.4cm}
\begin{equation}\label{eq:XORTick}
    t_{x} = w_c*(2*maxIter+5) + 2
\end{equation}
\normalsize
\vspace{-0.5cm}

We determine the coefficients on $maxIter$ for each design by the number of cores required to compute the VNU-CNU suggestions. The baseline implementation has an additional core required for the XOR calculation. The value added to this product is determined by the number of ticks needed in between words. One additional tick is needed for the XOR implementation to avoid interference of the last GaB iteration of the previous word with the first iteration of the next word. This expression represents the setup and overlapping execution for each word $w_c$, and finally the remaining constant is determined by remaining execution of the final word. These equations hold true for other H-Matrices as well, so long as the total core count does not change. We are now able to compute execution time on the ZCU102 with a given tick frequency for a sweeping range of $w_c$ and $maxIter$, which we use to estimate energy consumption as the design scales after we obtain power estimates.

\subsection{Resource Utilization and Power Analysis}

The addition of the xor and multiplexer in the neuron block along with control signals incurs a computational overhead, which we quantify in terms of resource utilization and  power estimation on the ZCU102. When gathering resource utilization and power estimates, we observe that the total length of execution is not a factor in these estimates. Thus, we use a subset of our initial dataset to gather these results. 

\begin{figure}[t]
    \centering
    \includegraphics[width=0.98\linewidth]{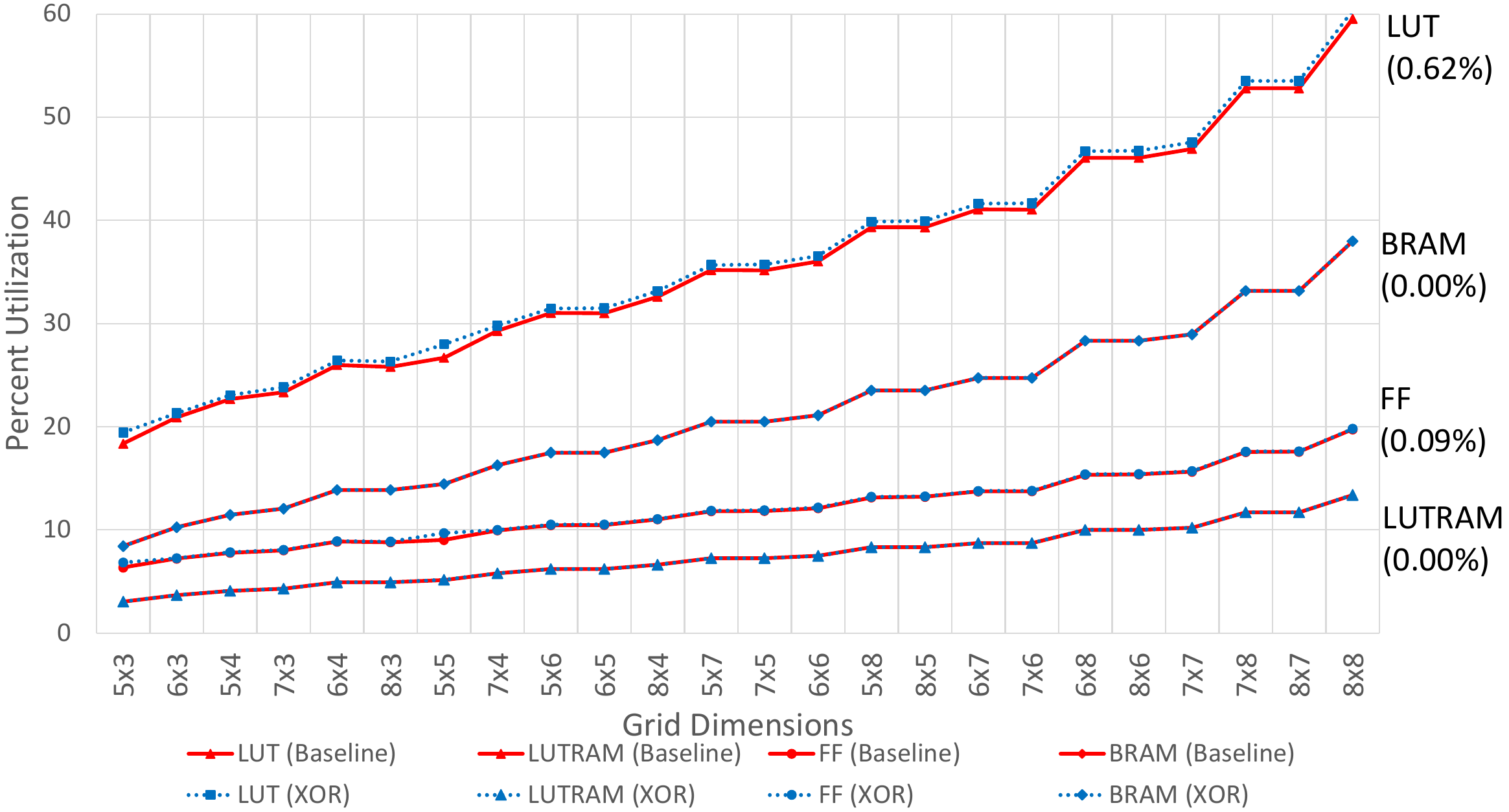}
    \caption{Resource utilization (\%)  of the baseline and XOR-integrated architectures with respect to grid size labeled as LUT, BRAM, FF and LUTRAM in RANC emulated on the ZCU102 evaluation board. The average difference for the XOR-architecture with respect to the baseline architecture across 24 grid dimensions are shown in parenthesis below the label.}
    \label{fig:resource_scaling}
    \vspace{-1mm}
\end{figure}

Figure~\ref{fig:resource_scaling} shows resource utilization ratio for each resource type on the FPGA with respect to the number of cores across 24 configurations. We observe that increasing the core count shows a trend where the XOR integrated design is at or slightly above the baseline by 0.62\% on average across the sweeping range in terms of LUT usage. For the other types (LUTRAM, FF, BRAM), the resource usage across all configurations show negligible difference between the two designs. Table~\ref{tab:power_comparison} shows the power estimation of the baseline and XOR-integrated designs on 5x3 and 5x5 grid dimensions. Similarly, the gap between the designs increases with the grid size (1.74\% on 5x5 and 1.20\% on 5x3).

\begin{figure*}[ht]
    \begin{subfigure}[c]{0.31\textwidth}
    \centering
    \scalebox{0.72}{
    \includegraphics[width=\linewidth, trim={0cm 0.5cm 0cm 1.5cm}, clip=true]{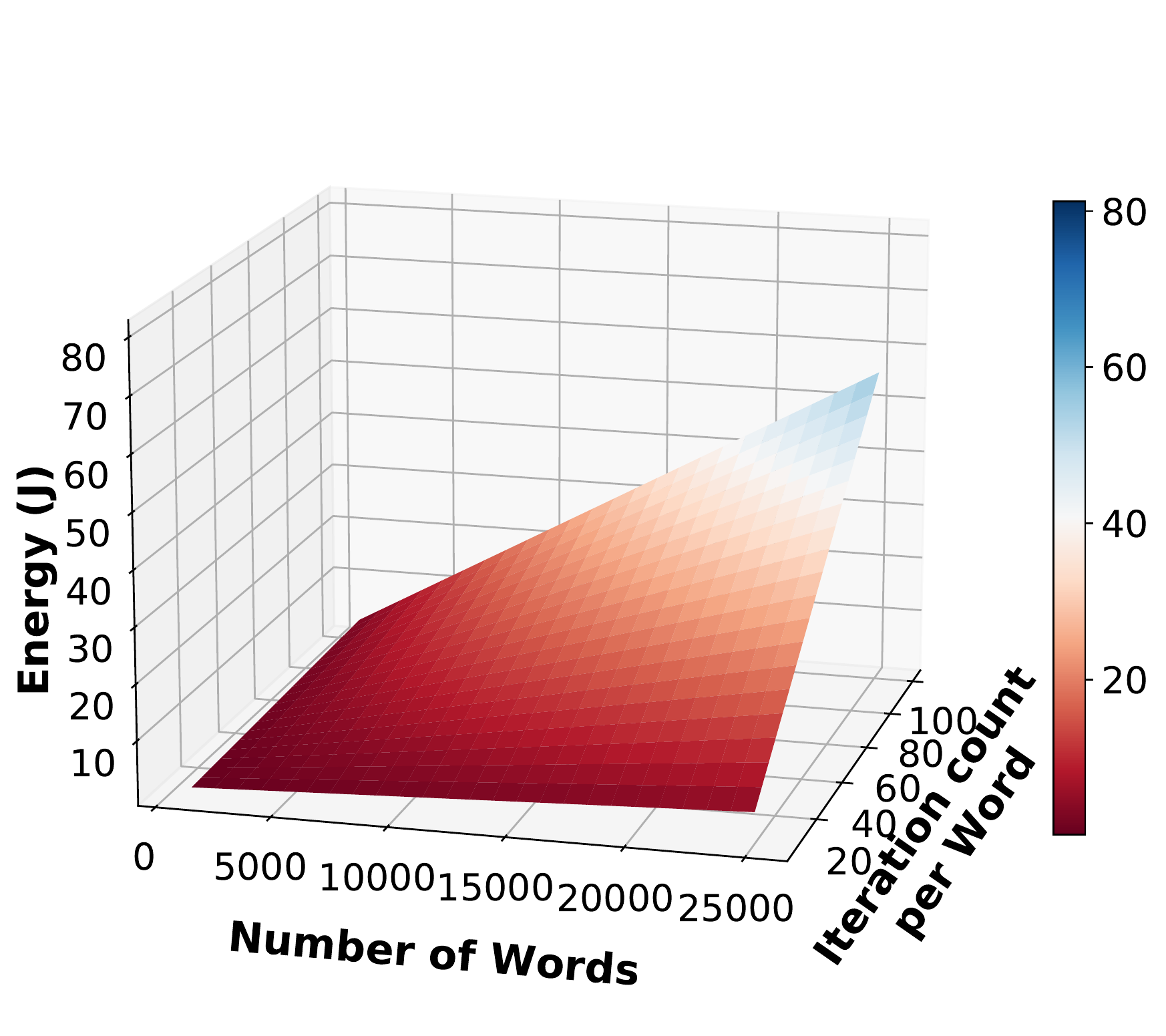}}
    \caption{XOR}\label{fig:energy_xor}
    \end{subfigure}
    \begin{subfigure}[c]{0.31\textwidth}
    \centering
    \scalebox{0.72}{
    \includegraphics[width=\linewidth, trim={0cm 0.5cm 0cm 1.5cm}, clip=true]{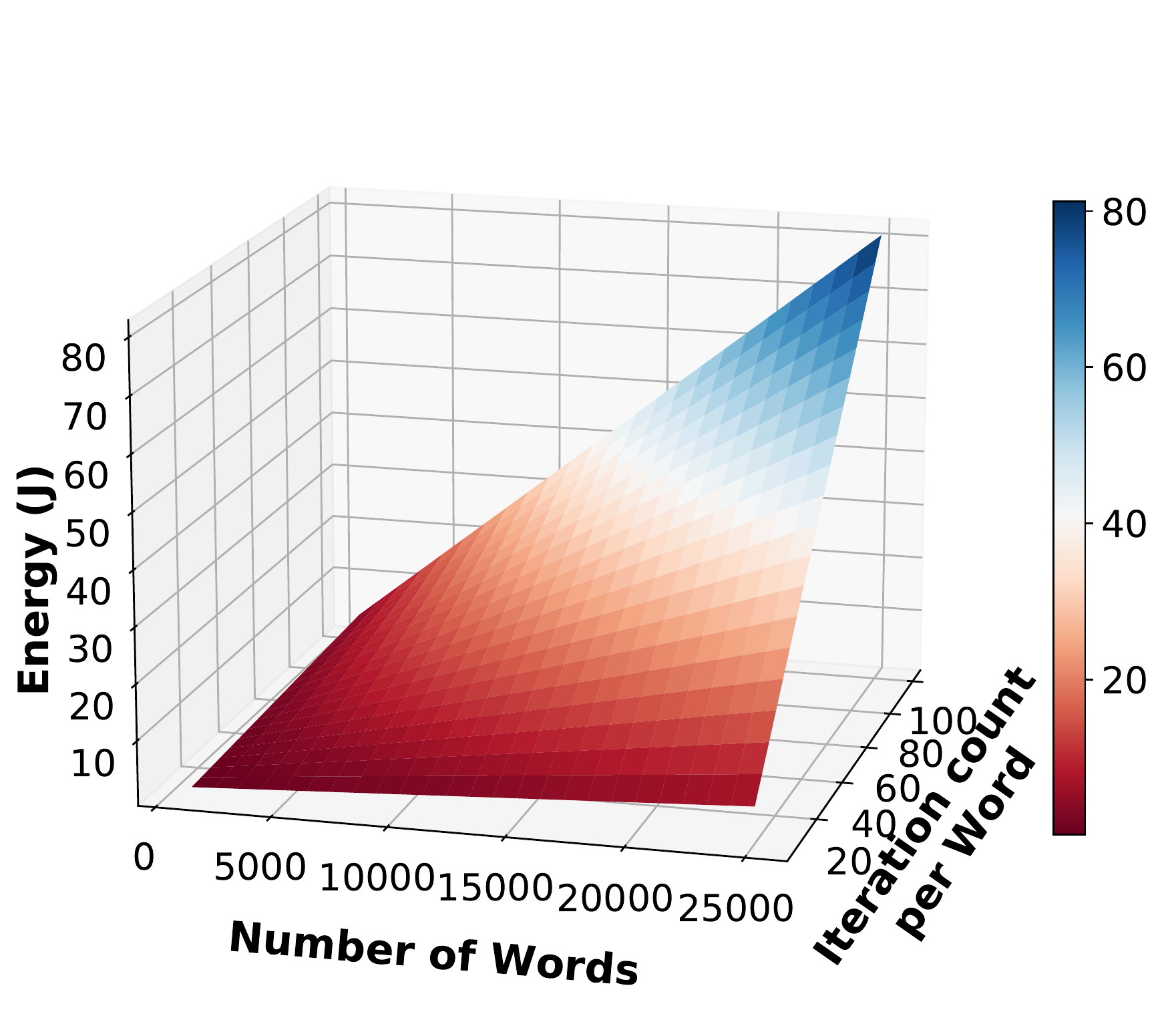}}
    \caption{Baseline}\label{fig:energy_baseline}
    \end{subfigure}
    \begin{subfigure}[c]{0.31\textwidth}
    \centering
    \scalebox{0.72}
        {
        \includegraphics[width=\linewidth, trim={0cm 0cm 0cm 2cm}, clip=true]{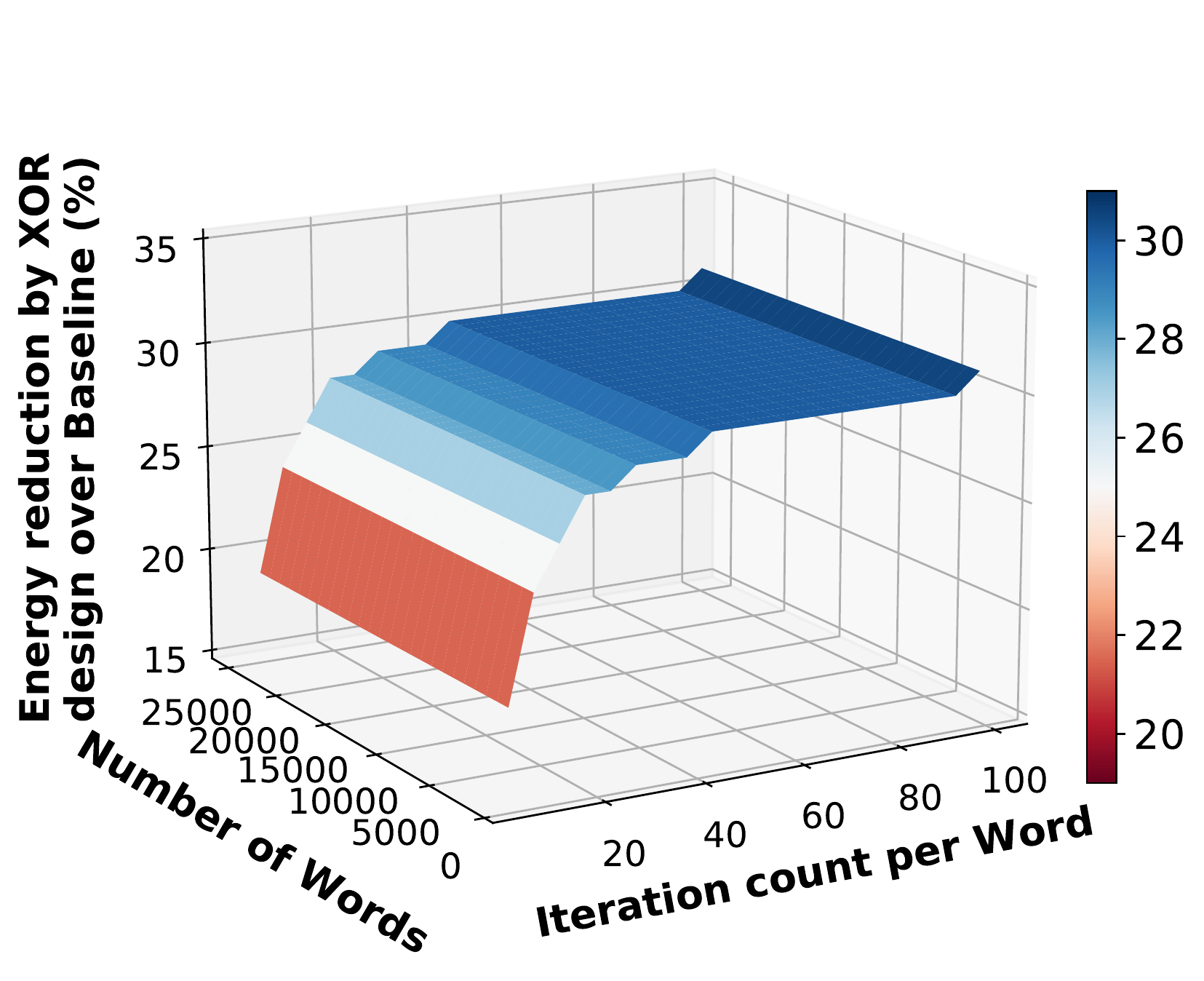}}
    \caption{XOR vs Baseline}
    \label{fig:energy_reduction_graph}
    \end{subfigure}
    
    \caption{Energy consumption of the (a) XOR-integrated and (b) baseline designs; (c) energy consumption reduction percentage by XOR-integrated design over the Baseline, on a 5x5 grid size with respect to number of words, and number of iterations per word. }\label{fig:energy_graphs}
    \vspace{-4mm}
\end{figure*}

Similar to the resource utilization, power estimates show for both grid sizes that the XOR-integrated design incurs a minor increase in power consumption over the baseline design. The 5x5 grid uses approximately 68\% more total power than the 5x3 grid for each design, however there is only 0.5\% more power consumed per core. The increase in total power is expected, as the FPGA is still allocating resources and running computations for each core, regardless of its participation in the GaB decoding algorithm, but individually, the cores are consuming negligibly more power for a given design when the grid size increases. This is important when considering the scaling of a design to support larger GaB decoders, since more cores will be needed to distribute computation for VNUs and CNUs. Additionally, we see that the difference in power and resource utilization between both designs does not scale when we increase the grid dimensions. The \% increase in both resource utilization and power between the designs remains largely consistent across grid dimensions. This suggests that if we scale the grid dimensions further to support larger decoders, we do not expect a larger \% increase in resource utilization or power in the FPGA environment. Further, we note from Figure~\ref{fig:ldpcxorgraph} in Section~\ref{subsec:xor} that the number of neurons for the baseline design increases quadratically as the CNU degree increases, instead of linearly as with the XOR-integrated design. Assuming there is a physical limit to the number of neurons in a core, this means that the grid sizes for the baseline design will also scale faster than the grid sizes for the XOR-integrated design. This will result in a worse overall power and resource utilization for the baseline design over the XOR-integrated design, as we have already shown that the grid dimensions have a larger impact than the neuron block design alone. Although the XOR-integrated design requires more power and resources over the baseline implementation, as we scale to larger decoders, the grid dimensions do not scale at the same rate for each design, showing favorable results for the XOR-integrated design. This also has implications for energy as we discuss in the next section.

\begin{table}[t]
\caption{Baseline and XOR power estimations, along with \% increase in power by XOR over baseline architecture, on the ZCU102 evaluation board.}
\vspace{-2mm}
\label{tab:power_comparison}
\tabcolsep=0.18cm
\centering
\scalebox{\tabsize}{
\begin{tabular}{|l|l|l|l||l|l|l|}
\cline{2-7}
\multicolumn{1}{c|}{} & \multicolumn{3}{c||}{\textbf{5x5 grid size}} & \multicolumn{3}{c|}{\textbf{5x3 grid size}} \\ \hline
\multirow{2}{*}{\textbf{Metric}} & \textbf{Baseline} & \textbf{XOR} &\textbf{\%} & \textbf{Baseline} & \textbf{XOR} & \textbf{\%} \\ &
\textbf{(mW)} & \textbf{(mW)} & \textbf{Inc.} & \textbf{(mW)} & \textbf{(mW)} & \textbf{Inc.} \\
\hline
Total Power & 656.875 & 668.300 & 1.74\% & 392.240 & 396.960 & 1.20\% \\ \hline
Core Power & 26.275 & 26.732 & 1.74\% & 26.149 & 26.464 &   1.20\% \\ \hline
\end{tabular}
}
\end{table}

\begin{table}[t]
\caption{Baseline and XOR-integrated energy estimates on the 288 word dataset using ZCU102 FPGA.}
\vspace{-2mm}
\label{tab:energy_comp}
\tabcolsep=0.18cm
\centering
\scalebox{\tabsize}{
\begin{tabular}{|l|l|l|l||l|l|l|}
\cline{2-7}
\multicolumn{1}{c|}{} & \multicolumn{3}{c||}{\textbf{5x5 grid size}} & \multicolumn{3}{c|}{\textbf{5x3 grid size}} \\ \hline
\multirow{2}{*}{\textbf{Metric}} & \textbf{Baseline} & \textbf{XOR} &\textbf{\%} & \textbf{Baseline} & \textbf{XOR} & \textbf{\%} \\ &
\textbf{(mJ)} & \textbf{(mJ)} & \textbf{Red.} & \textbf{(mJ)} & \textbf{(mJ)} & \textbf{Red.} \\
\hline
Total Energy & 870.75 & 597.391 & 31.4\% & 519.953 & 354.841 & 31.4\%\\ \hline
Core Energy & 34.830 & 23.896 & 31.76\% & 34.664 & 23.656 & 31.76\% \\ \hline
\end{tabular}
}
\end{table}

\subsection{RANC Energy Analysis} 
In this subsection, we analyze the energy consumption of the RANC environment running GaB, including emulation overhead, on the ZCU102. To estimate energy, we combine the execution time of the 288 word dataset with the power estimates for each design on the ZCU102 as shown in Table~\ref{tab:energy_comp}. Overall, we observe that reduction in tick count with the XOR-integrated design is heavily reflected in the energy benefits of the design, and is negligibly compromised by the minor increase in resource usage and power consumption. We show an estimated energy reduction with the XOR-integrated design over the baseline design using our dataset of 31.4\% on the 5x5 grid size, and 31.76\% on the 5x3 grid size. These energy reduction values closely reflect the 32.57\% tick reduction presented earlier, however are slightly lower because of the minor power increase in the XOR-integrated design.

To show that this energy reduction is consistent for realistically larger datasets, we perform sweeps for each implemented design over the number of words, and the number of GaB iterations per word. This is computed based on the execution time following equations~\ref{eq:baselineTick} and~\ref{eq:XORTick}, and the overall design power. We present Figure~\ref{fig:energy_graphs} and Figure~\ref{fig:energy_reduction_graph}, which show total energy and energy reduction respectively over different word counts and GaB iterations per word. Figure~\ref{fig:energy_graphs} compares how scaling the word count and the number of GaB iterations per word impacts the energy consumption of all cores in the XOR-integrated (Figure~\ref{fig:energy_xor}) and baseline (Figure~\ref{fig:energy_baseline}) designs. We observe that the number of GaB iterations per word has the largest impact on energy consumption. Increasing the number of GaB iterations increases the tick count for each word, affecting the total tick count by a factor of the GaB iteration increase. However, increasing the number of words only increases the number of ticks by a constant amount. Similarly, in Figure~\ref{fig:energy_reduction_graph}, we observe that the number of GaB iterations per word has more impact on energy reduction than the number of words decoded. Overall, it shows that our energy savings saturates around 65 GaB iterations per word to a 31\% reduction with the XOR GaB design over baseline. We also point out that the number of GaB iterations is also a factor when determining the desired error correction capability for a design. Therefore, we can achieve the same error correction capability with less energy by using the XOR modified design when considering realistic datasets. Overall, when we consider scaling our presented GaB mapping technique to a larger H-matrix and longer codewords, the XOR-based neuron block would maintain a similar 31\% reduction in energy consumption over a baseline LIF neuron block.

\subsection{Limitations and Further Energy Analysis on Neuromorphic Implementation}
In this work our focus is on understanding the limitations of neuromorphic architectures; exposing the performance bottlenecks and quantifying the benefits of XOR-integrated architecture. In order to hone in on architectural features, we limited our investigation to a problem size that was practical (8-bit codeword length) to explain and discuss through tick-by-tick execution scenarios. While we benefit from the flexibility offered by the FPGA-based emulation (pre-silicon) of neuromorphic devices to perform architecture trade analysis, the RANC ecosystem does not offer an event-driven execution for direct comparison with ASIC designs.  Additionally, the considered 8-bit decoder design is not readily scalable to the conventional decoder sizes in the literature. In the current state of design, it is not feasible to accurately extrapolate the expected (post-silicon) hardware performance on the physical neuromorphic device.  Therefore, a conventional synthesis approach based on RANC is not applicable to fairly evaluate the energy efficiency of the XOR-integrated architecture with respect to the ASIC/FPGA implementations from the literature. However, to bridge the gap between pre- and post-silicon evaluation, “spike count” and “energy-per-spike” data have been utilized in simulation and emulation based studies~\cite{FrenkelMorphIC2019, martin2021eqspike}. This has been integrated into tools such as  KerasSpiking developed by Nengo~\cite{Bekolay2014Nengo}, which generates energy estimates of Loihi~\cite{ Davies2018Loihi} and SpiNNaker ~\cite{Painkras2012SpiNNaker} architectures. Following the same model, we executed GaB implementation through the RANC software simulation tool over the baseline and XOR-integrated architectures and collected the total number of spikes emitted across the cores using the same set of codewords discussed in our experimental setup. The core dimensionality for this experiment is irrelevant since only cores relevant to the computation generate spikes. Based on the reported 109 pJ  energy per spike for TrueNorth~\cite{MartiTrueNorthEstimations}, we are able to compare the energy-efficiency of the XOR-integrated architecture with respect to the baseline architecture.

\begin{table}[b]
\vspace{2mm}
\caption{Spike count based energy comparison over the 288 word dataset. }
\vspace{-2mm}
\label{tab:spike_energy}
\tabcolsep=0.18cm
\centering
\scalebox{\tabsize}{
\begin{tabular}{|l|l|l|}
\hline
\textbf{Metric} & \textbf{Baseline} & \textbf{XOR-integrated} \\
\hline
Total Spike Count & 282,024 & 100,008 \\
\hline
Total Estimated Energy (mJ) & 30.74 & 10.90 \\
\hline
\end{tabular}
}
\end{table}

\begin{table}[b]
\caption{FPGA (ZCU102) resource utilization comparison when implementing GaB with codeword length of 1296 bits, parity bits of 648  (code rate of 0.5)  on  8x8 and 16x16 core configurations of XOR-integrated and baseline architectures respectively. }
\vspace{-2mm}
\label{tab:1296_coreutil}
\tabcolsep=0.18cm
\centering
\scalebox{\tabsize}{
\begin{tabular}{|l|l|l|l|}
\hline
\textbf{Architecture} & \textbf{LUT} & \textbf{FF} & \textbf{BRAM} \\
\hline
XOR-integrated (8x8) & 165,341 & 108,509 & 346.5 \\
\hline
Baseline (16x16) & 735,353\textsuperscript{*} & 434,917\textsuperscript{*} & 907.50\textsuperscript{*} \\
\hline
\end{tabular}
}
\end{table}

Table~\ref{tab:spike_energy} shows that, when executing GaB on RANC over the full 8-bit codeword dataset, the XOR-integrated architecture would consume 10.90 mJ of energy, achieving an energy reduction factor over baseline of 2.8X. Following the scalability analysis presented in Figure~\ref{fig:resource_scaling}, we estimate that to implement GaB for a regular codeword of size 1296 and code rate of 0.5~\cite{unal_2018}, the XOR-integrated architecture would require 63 cores, with 256 axons and neurons per core. The 8x8 configuration (last data point in Figure~\ref{fig:resource_scaling}), reaches LUT utilization of around 60\% (165K LUTs) on the ZCU102 FPGA. On the other hand, the baseline architecture would require 243 cores, which is not feasible to implement on the same FPGA. Table~\ref{tab:1296_coreutil} shows the resource usage for XOR-integrated architecture collected after placement and routing, whereas for the baseline architecture, the values (marked by \textsuperscript{*}) are based on synthesis of the nearest square array of 16x16 cores before place and route on the FPGA. Therefore, we conclude that for resource constrained environments, XOR-integrated architecture is more favorable.

\section{Conclusion}
\label{sec:conclusion}
In this study we present a novel mapping methodology for hard-decision iterative decoding algorithms on a neuromorphic architecture. We further enhance the processing capability of this mapped architecture by introducing XOR-based execution in neuron blocks, which is a fundamental operation for the hard-decision bit-flipping class of error correction algorithms. We demonstrate the efficiency of the proposed implementation in terms of tick count and energy estimation. 

As future work, we aim to build tools to automate the architectural mapping process for a given arbitrary H-matrix and maximum iteration count. We believe that such automation will allow rapid decoder implementation for standards such as WiFi and 5G. This automation paired with flexible neuromorphic development environments such as RANC, will enable exploration of novel neuromorphic computing architectures and implementation methodologies for a broader range of hard-decision decoders in the trade space of error correction performance, energy efficiency and resource utilization. Furthermore, the proposed neuromorphic implementations of majority voting, XOR, AND, OR, and, NOR operators facilitate implementation of applications from domains such as cryptography and bioinformatics. In conclusion, our mapping methodology and XOR-integrated neuron block pave the way for further research into architectural modifications to make complex and accurate hard-decision decoders, as well as other XOR-based applications, practical for power-constrained systems through neuromorphic computing.

\bibliographystyle{ieeetr}
\bibliography{references.bib}

\vskip -3.0\baselineskip plus -1fil

\begin{IEEEbiography}[{\includegraphics[width=0.9in,height=0.9in,clip,keepaspectratio]{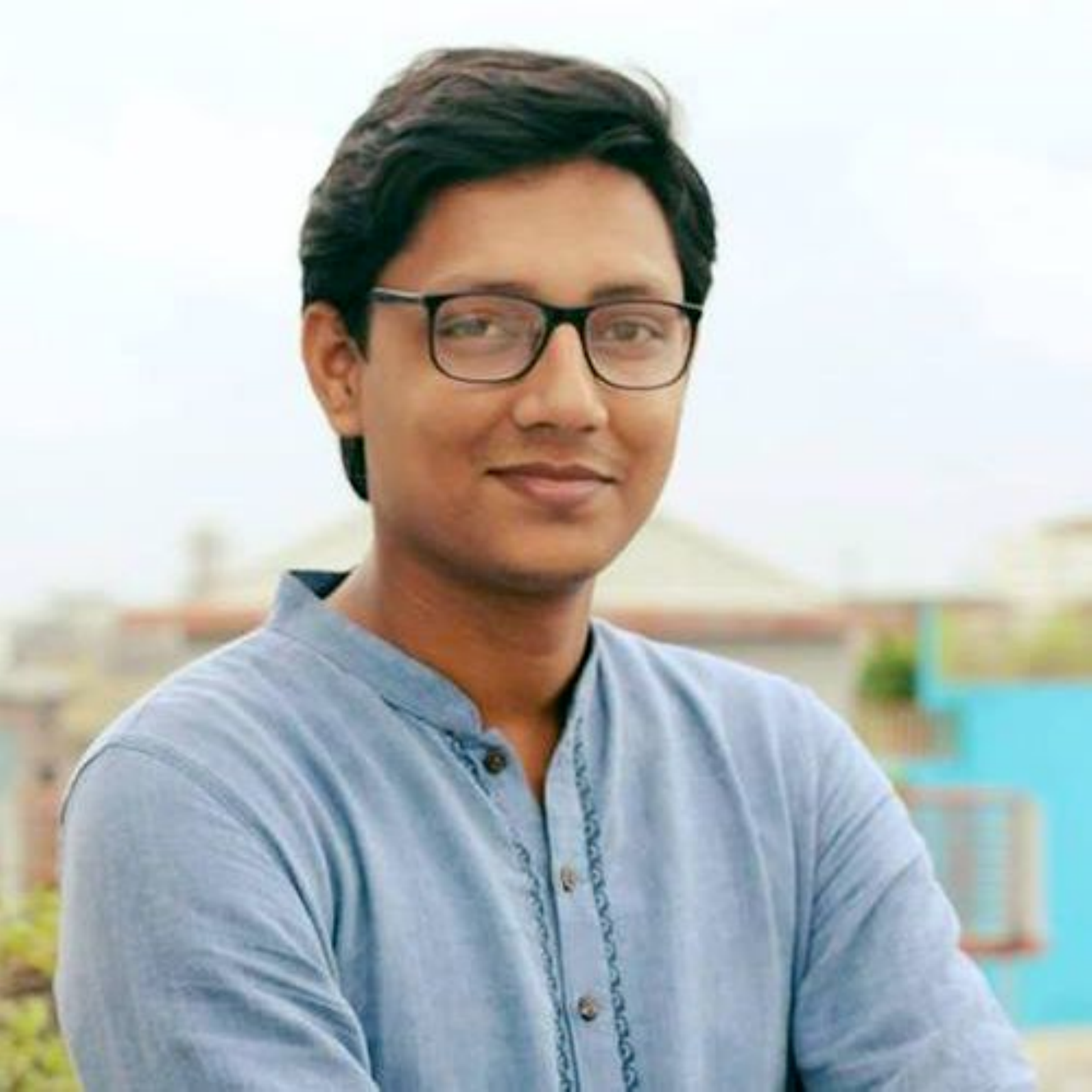}}]{Sahil Hassan}
is a Ph.D. student in the Electrical \& Computer Engineering program at the University of Arizona. He completed his B.Sc. and M.Sc. in Electrical and Electronic Engineering from University of Dhaka, Bangladesh. His research interests involve design of neuromorphic computing architectures, reconfigurable and heterogeneous computing systems, and adaptive hardware architectures.
\end{IEEEbiography}

\vskip -3.0\baselineskip plus -1fil

\begin{IEEEbiography}[{\includegraphics[width=0.9in,height=0.9in,clip,keepaspectratio]{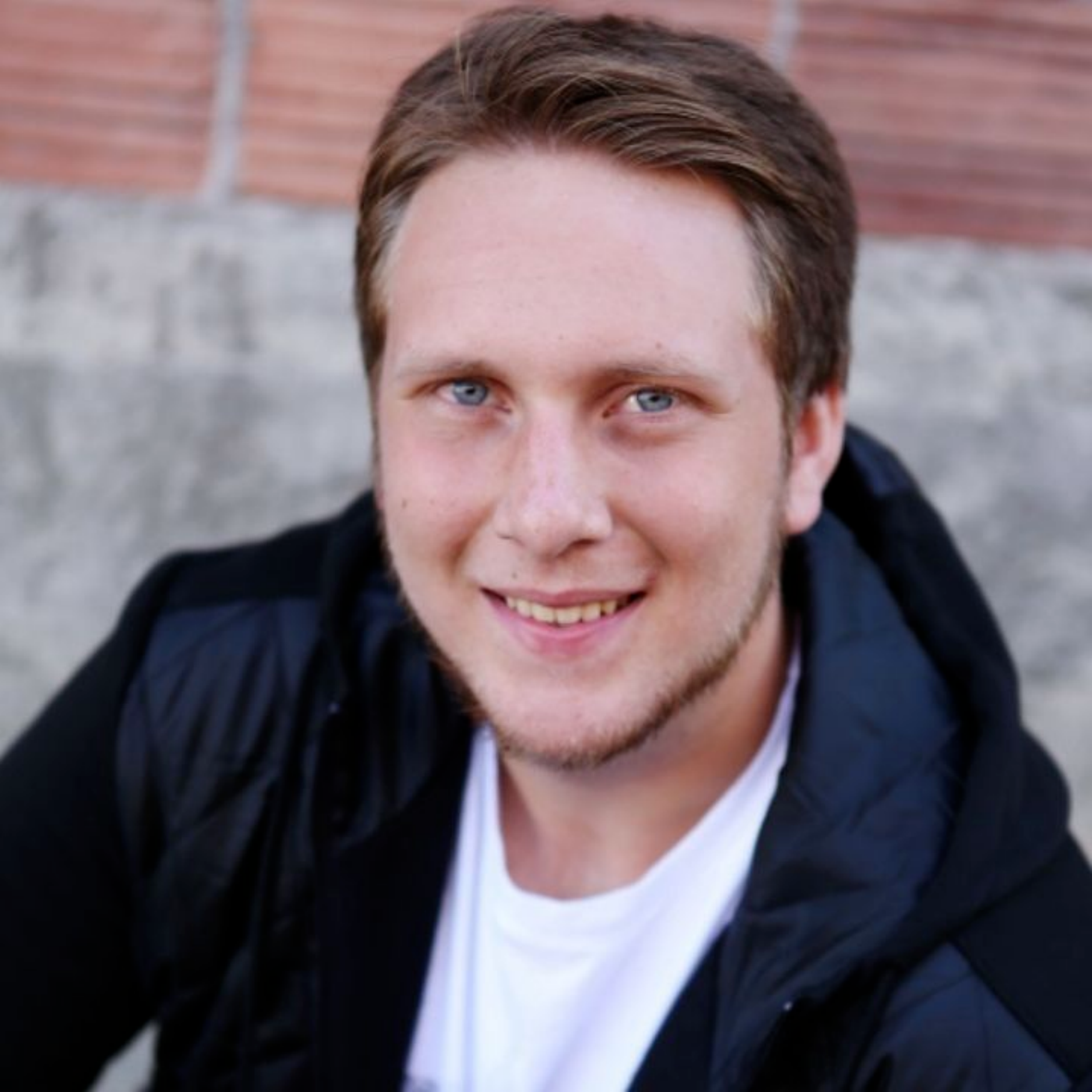}}]{Parker Dattilo} is an M.S. student in the Electrical \& Computer Engineering department at the University of Arizona. He also earned his B.S. degree in Electrical \& Computer Engineering at the University of Arizona. His research interests are design, architectural exploration, and hardware emulation of neuromorphic and reconfigurable computing systems. 
\end{IEEEbiography}

\vskip -3.0\baselineskip plus -1fil

\begin{IEEEbiography}[{\includegraphics[width=0.9in,height=0.9in,clip, keepaspectratio]{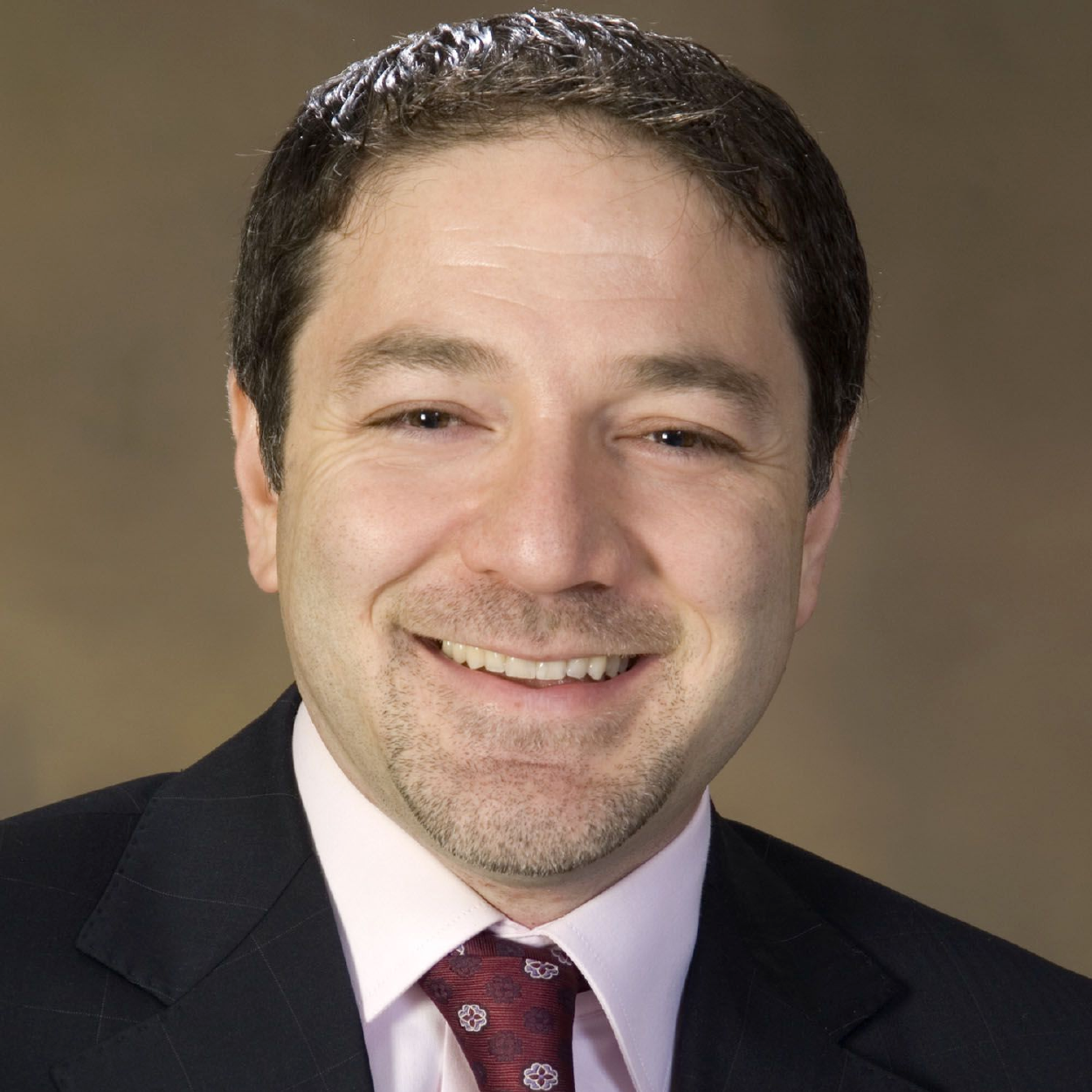}}]{Ali Akoglu}
received his Ph.D. degree in Computer Science from the Arizona State University in 2005. He is a Professor in the Department of Electrical \& Computer Engineering and BIO5 Institute at the University of Arizona. He is the site-director of the NSF Industry-University Cooperative Research Center on Cloud and Autonomic Computing. His research focus is on high performance computing and non-traditional computing architectures.
\end{IEEEbiography}

\end{document}